\documentclass{article}

\PassOptionsToPackage{numbers,sort&compress}{natbib}

\usepackage[preprint]{neurips_2025}

\usepackage[T1]{fontenc}
\usepackage[utf8]{inputenc}
\usepackage[english]{babel}
\addto\captionsenglish{%
}
\usepackage{microtype}
\UseMicrotypeSet[protrusion]{basicmath}  

\usepackage{amsmath,amsthm,amssymb,amsfonts,mathtools,thmtools}
\usepackage{bm}  
\usepackage{dsfont}  
\usepackage{nicefrac}  

\usepackage{graphicx}
\usepackage[font=footnotesize]{caption}  
\usepackage{adjustbox}  
\usepackage{booktabs}  
\usepackage{tabularx}
\usepackage{multirow}
\usepackage{wrapfig}
\usepackage[]{placeins}  

\usepackage[shortlabels]{enumitem}

\usepackage{algorithm}
\usepackage{algpseudocode}

\usepackage[svgnames]{xcolor}

\usepackage{seqsplit}  
\usepackage{fix-cm}  

\usepackage[unicode,bookmarksnumbered]{hyperref}  
\usepackage{xurl} 
\usepackage[capitalize,noabbrev,nameinlink]{cleveref}  

\newcommand{\muh}{\widehat{\mu}}
\newcommand{\Ch}{\widehat{C}}
\newcommand{\Qh}{\widehat{Q}}
\newcommand{\ph}{\widehat{p}}

\title{Improving Coverage in Combined Prediction Sets with Weighted p-values}

\author{
  Gina Wong\quad
  Drew Prinster\quad
  Suchi Saria\quad
  Rama Chellappa\quad
  Anqi Liu\\
  Johns Hopkins University
}

\begin{document}

\maketitle

\begin{abstract}
Conformal prediction quantifies the uncertainty of machine learning models by augmenting point predictions with valid prediction sets. For complex scenarios involving multiple trials, models, or data sources, conformal prediction sets can be aggregated to create a prediction set that captures the overall uncertainty, often improving precision. However, aggregating multiple prediction sets with individual $1-\alpha$ coverage inevitably weakens the overall guarantee, typically resulting in $1-2\alpha$ worst-case coverage. In this work, we propose a framework for the \textit{weighted aggregation of prediction sets}, where weights are assigned to each prediction set based on their contribution. Our framework offers flexible control over how the sets are aggregated, achieving tighter coverage bounds that interpolate between the $1-2\alpha$ guarantee of the combined models and the $1-\alpha$ guarantee of an individual model depending on the distribution of weights. Importantly, our framework generalizes to data-dependent weights, as we derive a procedure for weighted aggregation that maintains finite-sample validity even when the weights depend on the data. This extension makes our framework broadly applicable to settings where weights are learned, such as mixture-of-experts (MoE), and we demonstrate through experiments in the MoE setting that our methods achieve adaptive coverage.
\end{abstract}
\section{Introduction} \label{sec:intro}
\begin{figure*}[t]
\centering
\vspace{5pt}

\begin{minipage}[t]{0.5\textwidth}
    \vspace{0pt} 
    \centering
    \fontsize{8pt}{9pt}\selectfont{\textit{Example: storm forecasting}}
    \includegraphics[width=\linewidth]{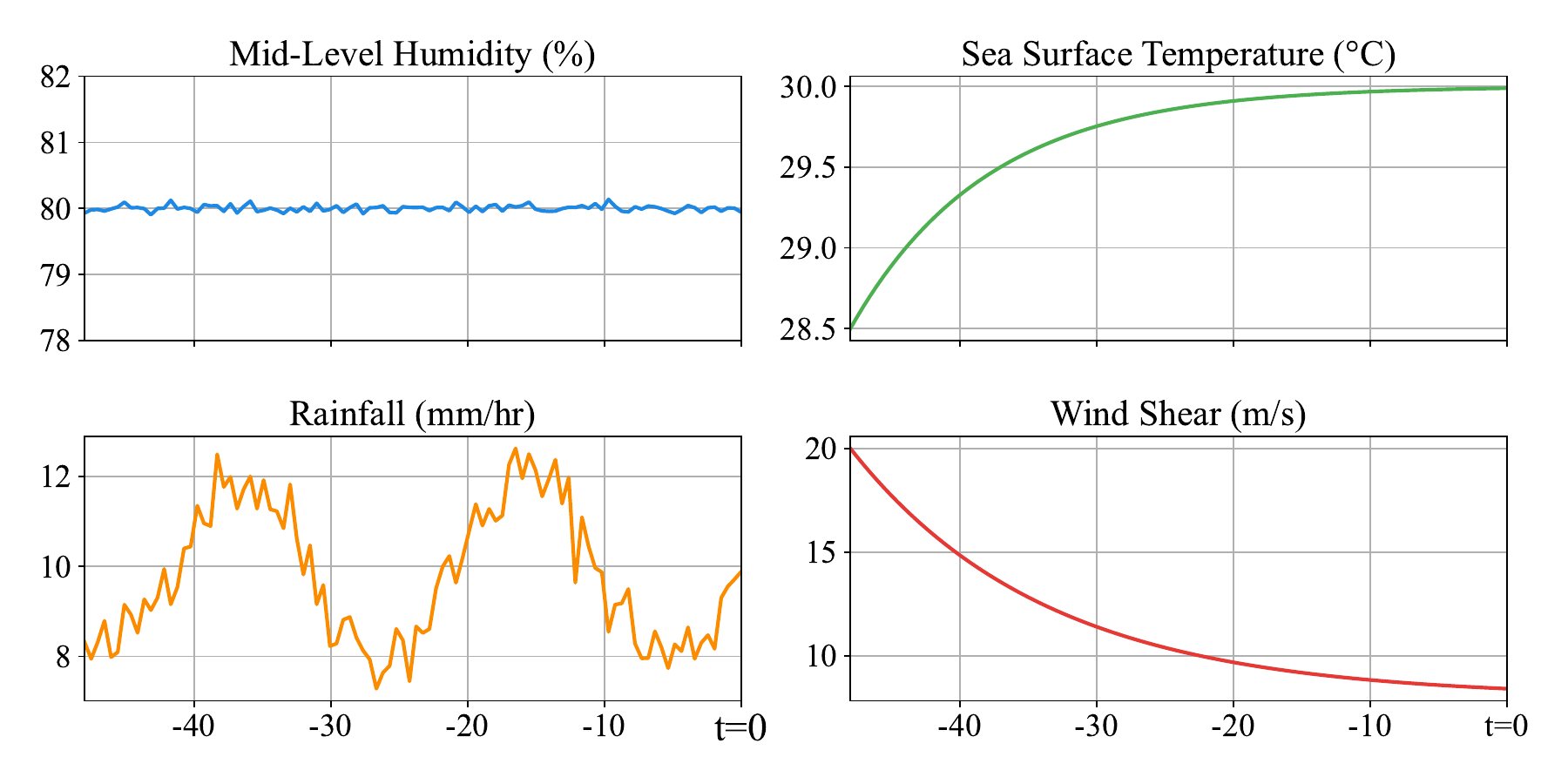}
    \includegraphics[width=0.95\linewidth]{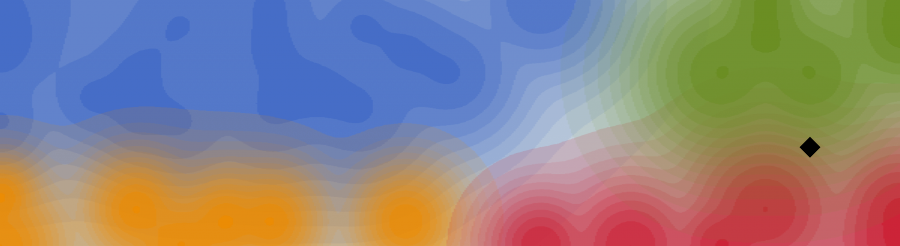}
\end{minipage} %
\hfill
\begin{minipage}[t]{0.48\textwidth}
    \vspace{0pt}  
    \raggedright
    \includegraphics[width=0.6\linewidth]{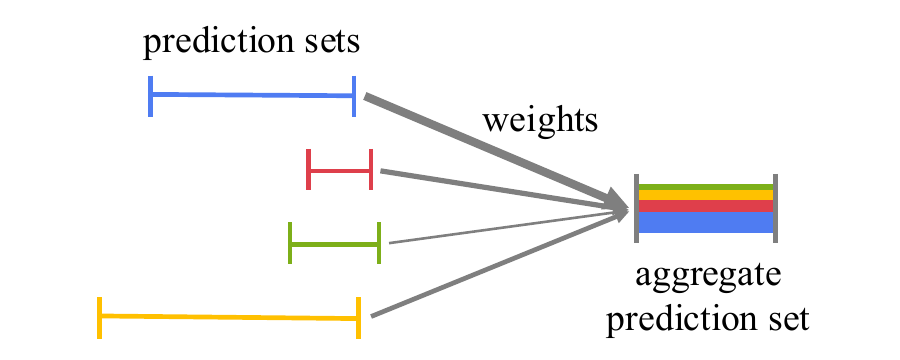} \\
    {\fontsize{7.8pt}{8.8pt}\selectfont
    (a) Weights are \textbf{data-independent} (\S \ref{sec:data_independent_weights})
    \begin{itemize}[leftmargin=2.5em,topsep=2pt,itemsep=0pt,parsep=0pt]
        \item tighter guarantee than previous work (Prop. \ref{prop:coverage_avg_constantw})
        \item expert priors: ``\textcolor{RoyalBlue}{humidity} is generally more reliable''
    \end{itemize}
    }
    \vspace{7pt}
    \includegraphics[width=0.6\linewidth]{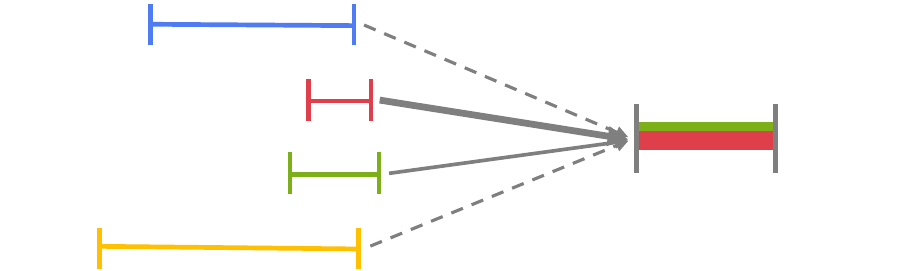} \\
    {\fontsize{7.8pt}{8.8pt}\selectfont
    (b) Weights are \textbf{data-dependent} (\S \ref{sec:data_dependent_weights})
    \begin{itemize}[leftmargin=2.5em,topsep=2pt,itemsep=0pt,parsep=0pt]
        \item adaptive coverage (e.g. mixture-of-experts \S \ref{sec:moe})
        \item \parbox[t]{\linewidth}{%
        personalized predictions: ``for $X$ location and $Y$ month,
        \textcolor{Crimson}{wind shear} is more important than
        \textcolor{OliveDrab}{sea temperature}''%
        }
    \end{itemize}
    } 
\end{minipage}
\caption{
\textbf{Left:} Storm forecasting example with different models tracking humidity, sea temperature, rainfall, and wind shear over time. Below, an abstract representation of how models vary in predictive strength across the input space (colored regions). At the given test point (black), the red and green models dominate, so their prediction sets matter most.
\textbf{Right:} Model prediction sets are combined with weighted aggregation.
(a) \textit{Data-independent} weights reflect expert priors (e.g.\ the blue model covers the largest portion of the input space and so is considered more generally reliable, and up-weighted accordingly).
(b) \textit{Data-dependent} weights adapt to context, yielding forecasts better aligned with current conditions (e.g.\ at the test point, red and green dominate).
\vspace{-10pt} 
}
\label{fig:summary_figure}
\end{figure*}

In recent years, machine learning models have achieved remarkable accuracy across a wide range of predictive tasks \citep{khan2022transformers, min2023recent, liang2024foundations}. Understanding the uncertainty associated with each prediction is essential for decision-making in real-world scenarios, but the black-box nature of many of these models hinders their deployment in safety-critical applications such as medical diagnosis \citep{richens2020improving, finlayson2021clinician, chua2023tackling, grote2023uncertainty}, industrial control systems \citep{kumar2023machine, lawrence2024machine}, and extreme weather forecasting \citep{kashinath2021physics, eyring2024pushing, lai2024machine}. Conformal prediction \citep{vovk2005algorithmic} emerged as a popular wrapper method around machine learning models because it provides a statistically valid quantification of uncertainty. In particular, it transforms point predictions to prediction sets with finite-sample coverage guarantees, as long as the test data is exchangeable with the training data. Complex scenarios with multiple predictions---for example, when there are multiple trials, models, or data sources---naturally produce multiple conformal prediction sets (Figure \ref{fig:summary_figure}).

A number of methods have been proposed to aggregate prediction sets. For conformal prediction, where the popular split conformal variant \citep{papadopoulos2002inductive} introduces a one-time random split, there are multiple methods of aggregating predictions to reduce the randomness over multiple splits: popular examples include cross-conformal \citep{vovk2015cross}, CV+ \citep{barber2021predictive}, and jackknife+ \citep{barber2021predictive}. These aggregation methods are all \textit{symmetric}, in that individual prediction sets contribute equally to the aggregate set. Although comparatively less studied, a more general approach is to aggregate sets \textit{asymmetrically}---that is, to weight sets based on their prior importance to the overall result \citep{gasparin2024merging, gasparin2024conformal}. Work in both symmetric and asymmetric aggregation establishes that aggregating individual prediction sets with $1-\alpha$ coverage guarantees results in an overall coverage guarantee of $1-2\alpha$ \citep{vovk2020combining}.

\newpage 

Our work is based on the observation that if the overall coverage guarantee for asymmetric aggregation reflects the contributions of the individual sets, then we can achieve a tighter guarantee than a constant $1-2\alpha$. Consider, as an extreme, the case where almost all the weight is on one set; then, because the coverage of that set dominates aggregation, the overall coverage guarantee should be closer to the $1-\alpha$ guarantee of the dominant individual set. To this end, we propose a method for asymmetric aggregation with \textit{data-independent weights} where the overall coverage guarantee is based on the distribution of weights across the prediction sets. This approach leverages the results of \citet{vovk2020combining} on averaging p-values to achieve tighter guarantees when there exists strong asymmetry in the importance of the prediction sets---i.e. when a single prediction set is significantly more important than the rest---with $1-2\alpha$ as the worst-case guarantee when the contributions of the prediction sets are more equal. 

Data-independent weights can encode expert priors on the relative importance of the prediction sets, allowing weights to reflect prior information such as the reliability of the predictors used to generate the sets (Figure \ref{fig:summary_figure}a). Already, these prior weights generalize standard symmetric aggregation. Yet in practice, many applications demand even more flexibility, and require weights that adapt directly to the data: for example, in the popular mixture-of-experts setting \citep{jacobs1991adaptive}, expert weights are learned from inputs. To deal with this general case, we additionally propose a method for aggregating prediction sets with \textit{data-dependent weights} via a linear transformation on the weighted average of p-values associated with each set. Our method allows us to construct a valid aggregate prediction set while also preserving the proportions of the weights assigned to the individual sets. Since data-dependent weights adjust the contributions of the individual sets based on the observed input, our method also achieves adaptive coverage (Figure \ref{fig:summary_figure}b). We apply our method to the mixture-of-experts setting, and we demonstrate its effectiveness in experiments with real and synthetic data.

To summarize, our main contributions are as follows:
\begin{itemize}[leftmargin=15pt,topsep=1pt,itemsep=0mm]
    \item We propose a framework for the asymmetric aggregation of prediction sets based on weighted p-values (\S \ref{sec:data_independent_weights}). With \textit{data-independent weights}, our framework improves the coverage guarantee beyond the standard $1-2\alpha$, depending on the distribution of weights (Prop. \ref{prop:coverage_avg_constantw}).
    
    \item We derive a general method for transforming a random variable to a p-value variable, which enables the construction of a prediction set with finite-sample guarantees (\S \ref{sec:data_dependent_weights}). This method allows us to extend our framework to \textit{data-dependent weights}, and we demonstrate that data-dependent weights provide adaptive coverage (Prop. \ref{prop:coverage_avg_dataw}, Prop. \ref{prop:coverage_avg_dataw_empirical}).
        
    \item We apply our method to the mixture-of-experts setting, and we demonstrate that using a weighted aggregation of experts improves local validity (\S \ref{sec:moe}).
\end{itemize}

\section{Background}
We begin by reviewing preliminaries leading up to the basics of conformal prediction, with a focus on the widely-used split conformal method. To provide context for our method based on p-values, we establish the connection between the more typical quantile presentation of conformal prediction and its original p-value presentation \citep{vovk2005algorithmic}.

\paragraph{Quantile and the empirical quantile function}
Given a random variable $X \in \mathcal{X}$, the left $\alpha$-quantile of $X$, which we refer to as the \textit{$\alpha$-quantile} of $X$ for simplicity, is defined as
\begin{equation*}
    q_\alpha(X) = \inf \{ x \in \mathcal{X} : \mathbb{P}(X \leq x) \geq \alpha \}.
\end{equation*}
We denote the distribution function of $X$ as $F$. The $\alpha$-quantile is then the inverse of $F$; that is, $q_\alpha(X) = F^{-1}(\alpha)$.

For a set of $n$ elements $Z = \{z_1, \dots, z_n\}$, the left $\alpha$-\textit{empirical quantile} is given by
\begin{equation*}
    \Qh^-_\alpha (Z) = \text{the } \left\lfloor (n+1)\alpha \right\rfloor \text{-th smallest value of } Z,
\end{equation*}
and the right $\alpha$-empirical quantile, or the $(1-\alpha)$-empirical quantile, by
\begin{equation*}
\Qh^+_\alpha (Z) = \text{the } \left\lceil (n+1)(1-\alpha) \right\rceil \text{-th smallest value of } Z = - \Qh^-_\alpha ( -Z ),
\end{equation*}
where $-Z = \{-z_1, \dots, -z_n\}$. Note that the $n+1$ term serves as a finite sample correction.

\paragraph{p-values}
A \textit{p-variable} \citep{vovk2020combining} is a random variable $P'$ such that
\begin{equation} \label{eq:p-variable}
    \mathbb{P} \{ P' \le \alpha \} \le \alpha \quad \forall \alpha \in (0,1).
\end{equation}
The values taken by a p-variable are called \textit{p-values}. In hypothesis testing, p-values represent the probability of observing results at least as extreme as the ones obtained, assuming the null hypothesis is true. When the inequality becomes equality, the p-variable is \textit{precise} (rather than  \textit{conservative}) and follows a $\mathcal{U}(0, 1)$ distribution.

\paragraph{Conformal prediction}
Suppose we have training data $\{(X_1, Y_1), \dots (X_n, Y_n)\}$ and a test point $(X_{n+1}, Y_{n+1})$ with unknown label $Y_{n+1}$. Let $\mathcal{P}$ denote the joint distribution on $(X,Y) \in \mathcal{X}\times\mathcal{Y}$. Then, assuming that the training and test data are \textit{exchangeable}, or that the distribution of the training and test data is permutation-invariant, conformal prediction can be used to construct a valid prediction set for $X_{n+1}$ with no further assumptions on $\mathcal{P}$.

Split conformal prediction, a popular variant, starts by splitting the training data indices into disjoint ``proper'' training and calibration subsets so that $\{1, \dots, n\} = S_{\text{train}} \cup S_{\text{cal}}$. We fit a predictor to the subset $S_{\text{train}}$, so that $\muh = \mathcal{A}(\{(X_i, Y_i)\}_{i\in S_{\text{train}}})$, where $\mathcal{A}$ is a model fitting algorithm. Then, given a nonconformity score function $s: \mathcal{X} \times \mathcal{Y} \rightarrow \mathbb{R}$, we can compute the scores $R_i = s(X_i, Y_i)$, or the values of the score function that characterize how nonconformal a label $Y_i$ is from its predicted value $\muh(X_i)$. We compute the scores on the subset $S_{\text{cal}}$. At a given significance level $\alpha$, this procedure allows us to define the prediction set
\begin{equation*}
\Ch_{\alpha}^{\text{split}}(X_{n+1}) = \left\{ y \in \mathcal{Y} : s(X_{n+1}, y) \le \Qh^+_\alpha(\{s(X_i, Y_i)\}_{i \in S_{\text{cal}}})\right\}.
\end{equation*}
By construction, the split conformal prediction set results in the (marginal) coverage guarantee
\begin{equation*}
    \mathbb{P} \{ Y_{n+1} \in \Ch^{\text{split}}_{\alpha}(X_{n+1}) \} \ge 1-\alpha \quad \forall \alpha \in (0,1).
\end{equation*}
In other words, the prediction set $\Ch^{\text{split}}_{\alpha}(X_{n+1})$ is \textit{valid}, or guaranteed to contain the true label $Y_{n+1}$ with probability at least $1-\alpha$. 

As an alternative perspective, the quantile can be replaced with the p-value function
\begin{equation*}
    \ph(x,y) = \frac{1 + \sum_{i \in S_{\text{cal}}} \mathds{1}\{s(x,y) < s(X_i, Y_i)\}}{|S_{\text{cal}}| + 1},
\end{equation*}
which returns the proportion of calibration points that are less conformal than some test point $(x,y)$, with a finite sample correction.
The equivalent prediction set constructed from the p-value function is 
\begin{equation*}
    \Ch_{\alpha}^{\text{split}}(X_{n+1}) = \left\{ y \in \mathcal{Y} : \ph(X_{n+1},y) > \alpha \right\}.
\end{equation*}

\section{Related work}
We categorize existing work on combining conformal results into two approaches: methods for selecting or combining prediction sets, and methods for combining p-values.

\textbf{Combining prediction sets~~}
Combining conformal prediction sets was first introduced by \citet{lei2018distribution}, who propose combining $K$ split conformal prediction sets by taking their intersection. If each prediction set is constructed at significance level $\alpha/K$, then a union bound yields an overall $\alpha$-level intersection set. However, because each individual prediction set targets the more stringent level $\alpha/K$, they tend to be inefficient---that is, their width or volume is large. Although taking the intersection of these sets reduces the size of the resulting prediction set, the authors show that, under general conditions, this intersection can still be wider than an $\alpha$-level split conformal prediction set, with probability tending to 1 with an asymptotic number of samples.

Rather than combine prediction sets, \citet{yang2024selection} propose selecting, from a set of $K$ predictors, the predictor that returns the best prediction set. They present two methods: one to select the predictor with the most efficient prediction set, but with only approximate validity, and one to select the predictor with the most valid prediction set (i.e.\ with minimal coverage slack), albeit with width only close to the minimum. \citet{liang2024conformal} point out that the approximate validity of the first method is due to selection bias, and introduce an alternative approach that both selects the most efficient predictor and uses it to construct a valid prediction set. Moving beyond choosing a single predictor, \citet{luoconformity} propose constructing an efficient and valid prediction set by a weighted combination of score functions in the multiclass classification case. In contrast to these methods, which commit to a single prediction set rule globally for all inputs, \citet{hegazy2025valid} propose a pointwise selection method: for each $X$, they (randomly) choose among candidate prediction sets in a stability-aware way that favors smaller sets, while preserving an appropriately adjusted marginal coverage guarantee.

A recent line of work explores combining conformal prediction sets by majority vote with a $1 - 2\alpha$ coverage guarantee \citep{cherubin2019majority, solari2022multi, gasparin2024merging, gasparin2024conformal}. \citet{gasparin2024merging} characterize the width of the majority vote set and introduce many extensions, including a weighted majority vote method that incorporates prior information in the weights. Our work on combining conformal results by weighted p-values is closely related to this method. However, our formulation allows us to leverage the results of \citet{vovk2020combining} to improve the coverage guarantee beyond the standard $1 - 2\alpha$, and can also be generalized to data-dependent weights to enable adaptive coverage. (See \S \ref{app:wmv} for more detail.)

\textbf{Combining p-values~~}
A substantial body of research has been devoted to developing methods for combining p-values. Here, we focus specifically on approaches that accommodate arbitrary dependence, with a particular emphasis on their applications to conformal prediction. (For a more comprehensive review, see \citet{balasubramanian2015conformal} and \citet{diciccio2020exact}.)

An early example of combining p-values with arbitrary dependence is the Bonferroni method, where the minimum of a set of p-values is scaled by the number of p-values. This method has multiple extensions \citep{ruger1978maximale, hommel1983tests}, and was first applied to conformal prediction by \citet{lei2018distribution}.

\citet{ruschendorf1982random} finds that twice the average of p-values is a p-value, a result that was later extended to a more general notion of average by \citet{vovk2020combining}. \citet{stutz2023conformal} use the result of \citet{ruschendorf1982random} to get a $1-2\alpha$ guarantee for the average of p-values, but also propose a novel transformation to get a p-value average with improved coverage. (For a discussion of how to extend their method to apply to our framework, see \S \ref{app:comparison_transformation_stutz}.)

\citet{liu2020cauchy} propose the Cauchy combination test, where the test statistic is a weighted sum of Cauchy-transformed p-values. \citet{wu2023multi} apply the Cauchy combination test to conformal prediction, and show that their method yields asymptotically exact prediction sets under a weak stability assumption and a joint distribution assumption.

\section{Combining conformal prediction sets by weighted p-value} \label{sec:data_independent_weights}
In this section, we extend the method of combining p-values, as proposed by \citet{vovk2020combining}, to the setting of conformal prediction. Suppose we have $K$ predictors, denoted by $\muh_1, \dots, \muh_K$, where each predictor $\muh_k$ is associated with a nonconformity score function $s_k$ and calibration set $S_k$. For each predictor, we define a p-value function $ \ph_k(x, y) $ that measures how well a candidate label $y$ conforms to the predicted outcome for a given $x$:
\begin{equation} \label{eq:pk}
\ph_k(x, y) = \frac{1 + \sum_{i \in S_k} \mathds{1}\{s_k(x, y) < s_k(X_i, Y_i)\}}{|S_k| + 1}.
\end{equation}
The corresponding prediction set for $X_{n+1}$ is 
\begin{equation*}
\Ch_{\alpha,k}^{\text{split}}(X_{n+1}) = \{y \in \mathcal{Y} : \ph_k(X_{n+1}, y) > \alpha\},
\end{equation*}
with marginal coverage guarantee
\begin{equation} \label{eq:split_coverage}
\mathbb{P} \left\{ Y_{n+1} \in \Ch_{\alpha,k}^{\text{split}}(X_{n+1}) \right\} \geq 1 - \alpha \quad \forall \alpha \in (0,1).
\end{equation}
We assign a weight $v_k$ to each p-value function $\ph_k$, yielding the weighted average p-value function
\begin{equation} \label{eq:p_avg}
    \bar{p}(x, y) = \sum_{k=1}^K v_k \ph_k(x, y) \quad\text{where } \sum_{k=1}^K v_k = 1,
\end{equation}
with prediction set
\begin{equation} \label{eq:ps_avg}
\Ch_{\alpha}^{\text{avg}}(X_{n+1}) = \{y \in \mathcal{Y} : \bar{p}(X_{n+1}, y) > \alpha\}.    
\end{equation}
We now provide a coverage guarantee for this aggregated prediction set.
\begin{restatable}{proposition}{CoverageConstantW}
\label{prop:coverage_avg_constantw}
Let $\Ch_{\alpha,1}^{\text{split}}(X_{n+1}), \dots, \Ch_{\alpha,K}^{\text{split}}(X_{n+1})$ be $K$ prediction sets defined by p-value functions $\widehat{p}_1, \dots, \widehat{p}_K$ \eqref{eq:pk} on $X_{n+1}$, where $1-\alpha$ coverage \eqref{eq:split_coverage} holds for each set $k \in [K]$. Then, the prediction set $\Ch_{\alpha}^{\text{avg}}(X_{n+1})$ \eqref{eq:ps_avg} from thresholding the weighted average p-value function \eqref{eq:p_avg} gives the coverage guarantee
\begin{equation*}
\mathbb{P} \left\{ Y_{n+1} \in \Ch^{\text{avg}}_{\alpha}(X_{n+1}) \right\} \ge 1 - \min \left\{ \frac{1}{v}, 2 \right\} \alpha \quad \forall \alpha \in (0,1),
\end{equation*}
where $v = \max\{v_1, v_2, \dots, v_K\}$ is the largest weight assigned to any of the p-values.
\end{restatable}

\paragraph{Weighted aggregation provides more flexible coverage guarantees}
In a sense, this result provides a spectrum of coverage guarantees based on the weight distribution among the models. When one predictor dominates (i.e., $v > 1/2$), the guarantee improves beyond the standard $1 - 2\alpha$ of typical aggregation methods, allowing the overall guarantee to approach that of the most influential predictor. In the extreme case, when all the weight is assigned to a single model ($v = 1$), we recover the standard split conformal guarantee of $1 - \alpha$ coverage for that model. Thus, we can interpolate between the standard $1 - 2\alpha$ guarantee of combined models and the $1 - \alpha$ guarantee of individual models, with the weights controlling the trade-off. This improves the coverage guarantee for asymmetric aggregation, and also opens up the utility of asymmetric generalizations of the many established symmetric aggregation methods (e.g. set-weighted versions of cross-conformal, CV+, etc.)

Since the work of \citeauthor{vovk2020combining} holds for arbitrarily dependent p-values, this method is robust across a wide range of scenarios. In practice, independent weights allow users to incorporate prior knowledge about the relative quality of different predictors \citep{vovk2020combining, gasparin2024merging}. 
The weighting can reflect, for example, expert insights on which of $K$ models should be prioritized as being more reliable.

\paragraph{From p-values to prediction sets~~}  
The work of \citeauthor{vovk2020combining} is central to our result, allowing us to generalize and improve upon existing conformal guarantees. Still, despite its broad applicability, their work remains underexplored in conformal literature, and the weight-dependent coverage result has not yet been applied to conformal prediction sets.\footnote{To the best of our knowledge, the works of \citeauthor{gasparin2024merging}~\citep{gasparin2024merging,gasparin2024conformal} are the only ones so far to use asymmetric set aggregation, and they derive the standard $1-2\alpha$ guarantee of symmetric aggregation methods.} We attribute this oversight to several factors.

\citeauthor{vovk2020combining} frame their work in terms of merging functions and p-values, without reference to prediction sets or conformal prediction. As a result, subsequent research has similarly focused on p-values and related test statistics, and has limited connection to conformal literature. Meanwhile, conformal literature typically uses a quantile-based construction of prediction sets, rather than a p-value construction, and work in conformal prediction set aggregation typically operates directly on the sets, rather than working with the associated p-value functions. These trends may contribute to the pattern where p-value results are not always propagated to the wider conformal prediction community. Our result shows that the p-value presentation offers unique benefits to prediction set aggregation, and we hope that our work may encourage renewed interest in the connection between p-values and conformal prediction.

\section{Extending to data-dependent weights} \label{sec:data_dependent_weights}
In many practical settings, model weights are determined by the observed data rather than fixed in advance. Such data-dependent weights naturally adapt the influence of each model to the characteristics of the input, making them central to applications like ensemble learning and mixture-of-experts. This adaptivity, however, creates a technical challenge: once the weights depend on the data, they also depend on the associated p-variables, and the theory of \citet{vovk2020combining} no longer applies. Nevertheless, we can build on their approach to develop a method that allows us to recover a valid coverage guarantee, even when the weights are data-dependent.

The key idea is to directly use the definition of a p-variable \eqref{eq:p-variable}. The sum of weighted p-variables, where the weights are dependent on the p-variables, is not necessarily a p-variable. However, the weighted sum can \textit{become} a p-variable by a linear transformation that both satisfies the definition of a p-variable and preserves the proportions of the weights.

Let the p-variables of the predictors be $P_1, \dots, P_K \in \mathcal{U}$, where $\mathcal{U}$ is the set of all uniformly distributed random variables. The weights are given by a random vector $\mathbf{W} = (W_1, \dots, W_K)$ in the $(K-1)$-dimensional simplex $\Delta_{K-1} := \{\mathbf{w} = (w_1, \dots, w_K) \in [0,1]^K : w_1 + \dots + w_K = 1\}$ depending on the data $\{(X_i, Y_i)\}_{i\in [n]} \cup \{X_{n+1}\}$. We define the weighted average function
\begin{equation} \label{eq:p_all}
p_{\text{all}}(x,y; \mathbf{w}) = \sum_k w_k \widehat{p}_k(x,y),    \end{equation}
giving associated random variable $P_{\text{all}} := p_{\text{all}}(X,Y;\mathbf{W})$ with distribution $F_{P_{\text{all}}}$.
Then, we propose to learn the scalar
\begin{equation} \label{eq:m_correction_factor}
\begin{aligned}
    m^* &= \inf\{m \in \mathbb{R}^+ : \mathbb{P}\{ m P_{\text{all}} \le \alpha \} \le \alpha \quad \forall \alpha \in (0,1)\} \\
    &= \inf\{m \in \mathbb{R}^+ : F_{P_{\text{all}}}(\alpha/m) \le \alpha  \quad \forall \alpha \in (0,1)\} = \sup_{\delta \in (0, \infty)}\frac{F_{P_{\text{all}}}(\delta)}{\delta}.
\end{aligned}
\end{equation}
(See \S\ref{app:derivation_mstar} for a derivation of the supremum expression.)
The scaling defined in \eqref{eq:m_correction_factor} transforms the weighted average $P_{\text{all}}$ into a valid p-variable, recovering a coverage guarantee. We formalize this in the following result.

\begin{restatable}[Infinite-sample guarantee]{proposition}{CoverageDataW}
\label{prop:coverage_avg_dataw}
Let $\Ch_{\alpha,1}^{\text{split}}(X_{n+1}), \dots, \Ch_{\alpha,K}^{\text{split}}(X_{n+1})$ be $K$ prediction sets defined by p-value functions $\ph_1, \dots, \ph_K$ \eqref{eq:pk} on $X_{n+1}$ corresponding to p-variables $P_1, \dots, P_K$. Suppose $1-\alpha$ coverage \eqref{eq:split_coverage} holds for each set $k \in [K]$. Let $(W_1, \dots, W_K)$ be a random vector in $\Delta_{K-1}$ depending on $\{(X_i, Y_i)\}_{i\in[n]} \cup \{X_{n+1}\}$, and let $p_{\text{all}}$ be the weighted average \eqref{eq:p_all}, with random variable $P_{\text{all}}$.

The prediction set from thresholding $p_{\text{all}}$ is
$\Ch_{\alpha}^{\text{unscaled}}(X_{n+1}) = \{y : p_{\text{all}}(X_{n+1},y) > \alpha\}$,
and it satisfies the coverage guarantee
\begin{equation*}
\mathbb{P}\{Y_{n+1} \in \Ch^{\text{unscaled}}_{\alpha}(X_{n+1})\} \ge 1 - m^* \alpha \quad \forall \alpha \in (0,1).
\end{equation*}

The prediction set from thresholding $m^* p_{\text{all}}$ is
$\Ch_{\alpha}^{\text{scaled}}(X_{n+1}) = \{y : m^* p_{\text{all}}(X_{n+1},y) > \alpha\}$,
and it satisfies the coverage guarantee
\begin{equation*}
\mathbb{P}\{ Y_{n+1} \in \Ch^{\text{scaled}}_{\alpha}(X_{n+1}) \} \ge 1 - \alpha \quad \forall \alpha \in (0,1).
\end{equation*}
Here, $m^*$ is the correction factor defined in \eqref{eq:m_correction_factor}.
\end{restatable}

\begin{algorithm}[t]
\caption{Constructing a weighted aggregate of prediction sets with valid coverage}
\label{alg:full_procedure}
\begin{algorithmic}[1]
\State \textbf{Input:} Data with indices $\{1, \dots, n\} = S_{\text{cal}} \cup S_{\text{merge}}$, $K$ predictors from potentially overlapping datasets, $K$ weights which may depend on data (e.g.\ learned router weights), test example $X_{n+1}$.
\State \textbf{Output:} Prediction set around the test example $\Ch_{\alpha}^{\text{scaled}}(X_{n+1})$, with coverage of at least \mbox{$1 - (\alpha + \epsilon + \delta)$}.

\Statex
\State \textbf{Step 1.} Derive the p-value function $p_{\text{all}}$.
\State \quad 1.1. Using calibration set $S_{\text{cal}}$, derive p-value functions $\widehat{p}_k$ for each of the predictors \eqref{eq:pk}.
\State \quad 1.2. Compute the aggregated p-value function 
$p_{\text{all}} = \sum_{k=1}^K w_k \, \widehat{p}_k$, 
using the (potentially data-dependent) weights $w_k$.

\Statex
\State \textbf{Step 2.} Compute the correction factor $\widehat{m}^*$.
\State \quad 2.1. Using the points of merging set $S_{\text{merge}}$ with the function $p_{\text{all}}$, get samples of $\widehat{F}_{P_{\text{all}}}$.
\State \quad 2.2. Derive the correction factor $\widehat{m}^*$ from the samples of $\widehat{F}_{P_{\text{all}}}$ \eqref{eq:empirical_m_correction_factor}.

\Statex
\State \textbf{Step 3.} For test example $X_{n+1}$, construct the prediction set
\begin{equation*}
\Ch^{\text{scaled}}_\alpha(X_{n+1}) = \{y: \widehat m^* p_{\text{all}}(X_{n+1},y) > \alpha\}.    
\end{equation*}
\end{algorithmic}
\end{algorithm}

\paragraph{Computing $m^*$ with a merging set} 
By definition, $m^*$ is the minimal scaling factor that makes $P_{\text{all}}$ a valid p-variable, and its value is determined by the CDF $F_{P_{\text{all}}}$. Because $F_{P_{\text{all}}}$ is a population quantity that is not accessible in practice, we cannot evaluate $m^*$ directly. Instead, we define a computable proxy $\widehat{m}^*$ using an empirical CDF constructed from a designated \textit{merging set} $S_{\text{merge}}$\footnote{So named because $S_{\text{merge}}$ is used to learn the correction that makes the merging function \citep{vovk2020combining} for the weighted average empirically precise.}.

Typically, the empirical CDF for $P_{\text{all}}$ computed from $S_{\text{merge}}$ is given by 
\begin{equation*}
\widehat{F}_{P_{\text{all}}}(\alpha) = \frac{\sum_{i \in S_{\text{merge}}} \mathds{1} \{ p_{\text{all}}(X_i, Y_i; \mathbf{W}^{(i)}) \leq \alpha \}}{|S_{\text{merge}}|},
\end{equation*}
but in practice we use the more conservative estimate
\begin{equation}\label{eq:cons_empirical_cdf}
\widehat{F}^{\text{cons}}_{P_{\text{all}}}(\alpha) = \frac{\mathds{1} \{ \scalebox{0.92}{$\min$}_{i \in S_{\text{merge}}} p_{\text{all}}(X_i, Y_i; \mathbf{W}^{(i)}) \leq \alpha \} + \sum_{i \in S_{\text{merge}}} \mathds{1} \{ p_{\text{all}}(X_i, Y_i; \mathbf{W}^{(i)}) \leq \alpha \}}{1 + |S_{\text{merge}}|},
\end{equation}
as it tends to yield more stable estimates in finite samples. The corresponding empirical correction factor is
\begin{equation}\label{eq:empirical_m_correction_factor}
\widehat{m}^* = \max_{i \in S_{\text{merge}}} \frac{\widehat{F}^{\text{cons}}_{P_{\text{all}}}(p_{\text{all}}(X_i, Y_i; \mathbf{W}^{(i)}))}{p_{\text{all}}(X_i, Y_i; \mathbf{W}^{(i)})},
\end{equation}
following \eqref{eq:m_correction_factor}. (For more detail on why \eqref{eq:empirical_m_correction_factor} is equivalent to \eqref{eq:m_correction_factor} for an empirical CDF, see \S\ref{app:computing_mhatstar}.) The construction of $\widehat{m}^*$ yields the following finite-sample guarantee:

\begin{restatable}[Finite-sample guarantee]{proposition}{CoverageDataWEmpirical}
\label{prop:coverage_avg_dataw_empirical}
Under the same assumptions as Proposition~\ref{prop:coverage_avg_dataw}, 
fix a user-chosen failure probability $\delta \in (0,1)$ and set 
\begin{equation*}
\varepsilon = \sqrt{\frac{\log(2/\delta)}{2 |S_{\text{merge}}|}}.
\end{equation*}

The prediction set from thresholding $p_{\text{all}}$ is
$\Ch_{\alpha}^{\text{unscaled}}(X_{n+1}) = \{y: p_{\text{all}}(X_{n+1},y) > \alpha\}$,
and it satisfies the coverage guarantee
\begin{equation*}
\mathbb{P}\{Y_{n+1} \in \Ch^{\text{unscaled}}_{\alpha}(X_{n+1})\} \ge 1 - \big(\alpha \mathbb{E}[\widehat{m}^*] + \varepsilon + \delta\big) 
\quad \forall \alpha \in (0,1).
\end{equation*}
The prediction set from thresholding $\widehat{m}^* p_{\text{all}}$ is
$\Ch_{\alpha}^{\text{scaled}}(X_{n+1}) 
= \{y : \widehat{m}^* p_{\text{all}}(X_{n+1},y) > \alpha\}$,
and it satisfies the coverage guarantee
\begin{equation*}
\mathbb{P}\{ Y_{n+1} \in \Ch^{\text{scaled}}_{\alpha}(X_{n+1}) \} \ge 1 - (\alpha + \varepsilon + \delta) \quad \forall \alpha \in (0,1).
\end{equation*}

Here, $\widehat{m}^*$ is the empirical correction factor defined in \eqref{eq:empirical_m_correction_factor}.
\end{restatable}

The two formulations in Propositions~\ref{prop:coverage_avg_dataw} and \ref{prop:coverage_avg_dataw_empirical} demonstrate the complementary roles of $m^*$. In the unscaled sets, $m^*$ scales the coverage lower bound in parallel with  Proposition~\ref{prop:coverage_avg_constantw}, while in the scaled sets it rescales the p-variable to yield a standard $1-\alpha$ style guarantee. For consistency with other conformal methods, we use the scaled formulation in our experiments.

An important practical question is how the quality of the correction $\widehat{m}^*$ depends on the merging set $S_{\text{merge}}$. In \S\ref{app:synthetic_exp_smerge}, we study this dependence by varying $|S_{\text{merge}}|$, and we show that even a modest number of samples ($<200$) is sufficient for accurate coverage. 

We summarize our full procedure in Algorithm~\ref{alg:full_procedure}.

\paragraph{Data-dependent weights give a form of conditional coverage} Data-dependent weights allow the influence of each model to be adjusted based on how well it performs for a specific data point. This approach enables the construction of aggregated prediction sets that are tailored to the characteristics of the given data, which can be viewed as a form of \textit{locally conditional coverage}. To be clear, true X-conditional coverage---where the coverage guarantee is conditioned on the current input---is impossible without additional distributional assumptions \citep{lei2014distribution, vovk2012conditional}. However, we demonstrate in \S\ref{sec:moe} that data-dependent weights allow us to create aggregated prediction sets with data-adaptive coverage, which can greatly improve conditional validity in practice.

\paragraph{Achieving tighter guarantees}
Our scaling correction factor $m^*$ provides a coverage guarantee that holds for all significance levels $\alpha \in (0,1)$. However, guarantees on coverage for all $\alpha$ can lead to overly conservative prediction sets, which may be unnecessarily restrictive in practice when we do not need guarantees for every possible significance level. Thus, we propose two alternatives: for a specific significance level $\alpha'$, we can learn a correction factor
\begin{equation} \label{eq:m_correction_factor_tighter}
    m^\dag = \inf\{m \in \mathbb{R}^+ : \mathbb{P}\{ m P_{\text{all}} \le \alpha \} \le \alpha \quad \forall \alpha \in (0,\alpha']\},
\end{equation}
or the even stricter
\begin{equation} \label{eq:m_correction_factor_exact}
    m^\ddag = \inf\{m \in \mathbb{R}^+ : \mathbb{P}\{ m P_{\text{all}} \le \alpha' \} \le \alpha'\}.
\end{equation}
(Note that \eqref{eq:m_correction_factor_tighter} is not without precedent: for the $1-2\alpha$ coverage guarantees of aggregation methods like cross-conformal and jackknife+, $\alpha'=0.5$ is the highest significance level of interest.)

\section{Application: mixture-of-experts} \label{sec:moe}
Mixture-of-experts (MoE) is a machine learning framework designed to combine the predictions of multiple specialized models, called experts \citep{jacobs1991adaptive}. Each expert in an MoE focuses on unique aspects of the problem by learning different representations of the input data. A central component of this framework is the routing network, which determines how to combine the experts' outputs. Specifically, for an input $x$, the routing network of a traditional (soft) MoE assigns weights $W_k(x)$ to each expert output $f_k(x)$, producing the final prediction as a weighted sum
\begin{equation} \label{eq:moe}
    f(x) = \sum_k W_k(x) f_k(x).
\end{equation}
By learning how to route different inputs to the most appropriate experts, the routing network implicitly conditions the model’s final prediction on the combination of experts that fits the given input. In this way, the routing network can be viewed as learning a form of conditional relationship between the input features and the expertise of each model.

\begin{figure*}[t] 
\centering

\begin{minipage}{0.44\textwidth}
    \includegraphics[width=\textwidth]{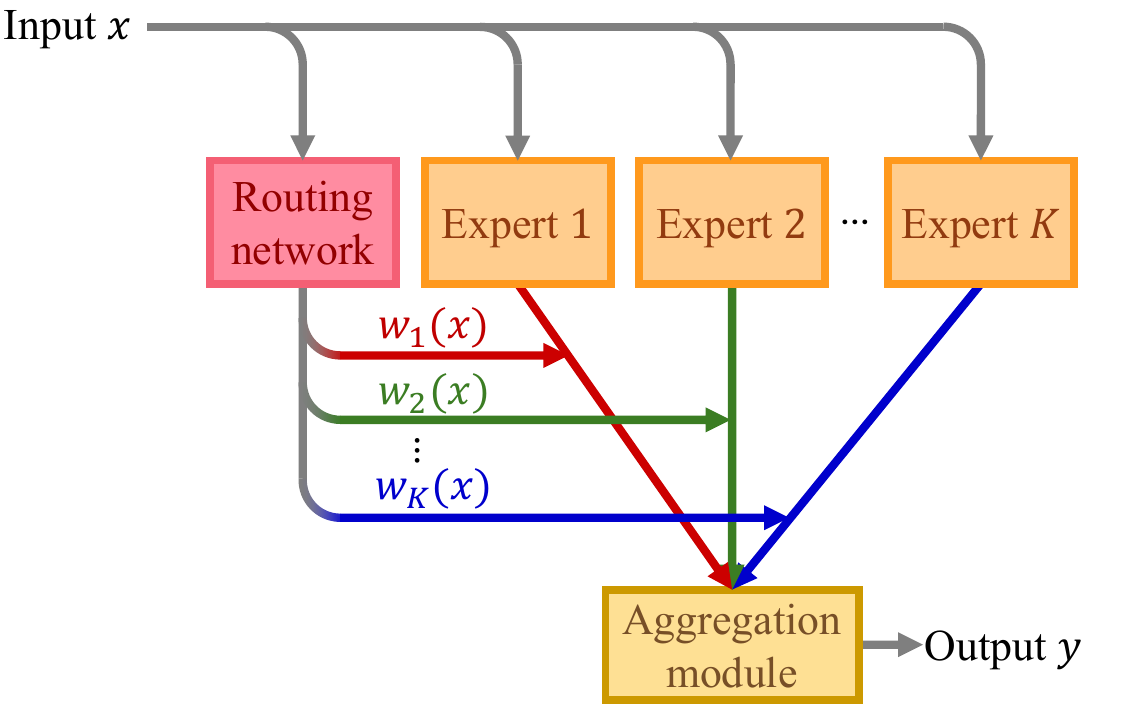}
\end{minipage}
\hfill
\begin{minipage}{0.55\textwidth}
    \includegraphics[width=0.49\textwidth]{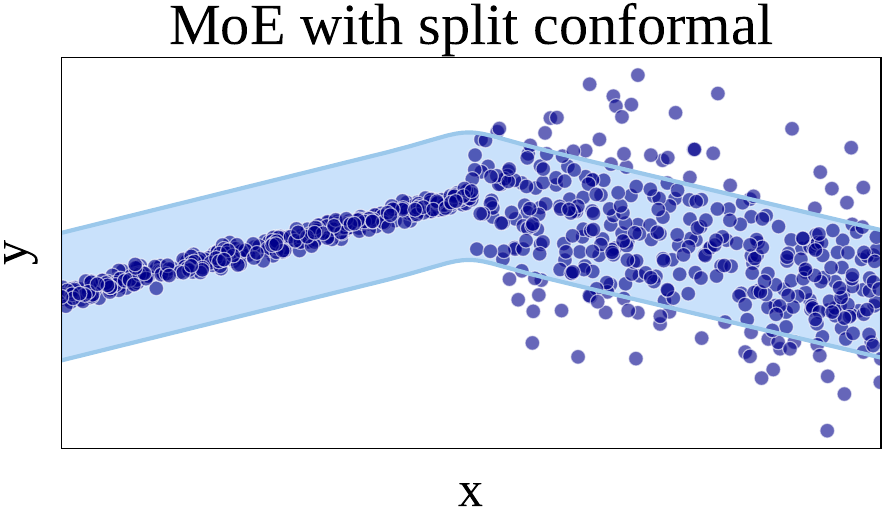}
    \hfill
    \includegraphics[width=0.49\textwidth]{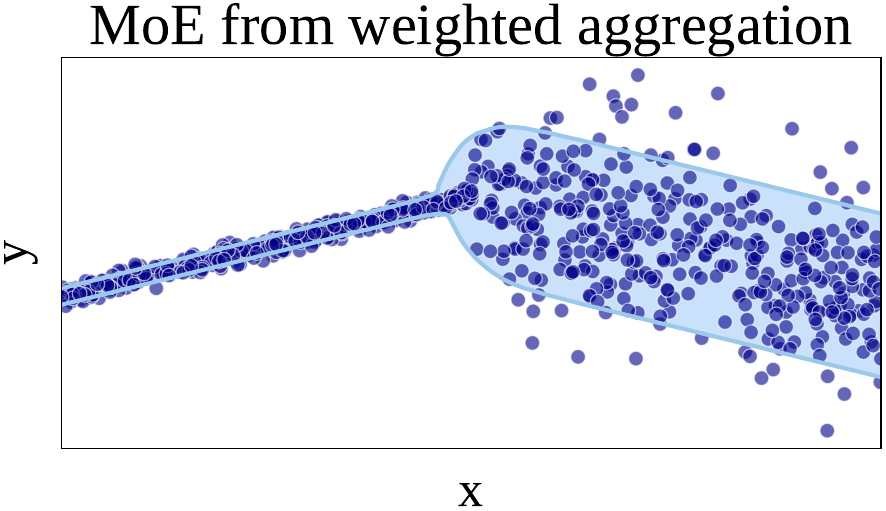} \\
    \includegraphics[width=0.49\textwidth]{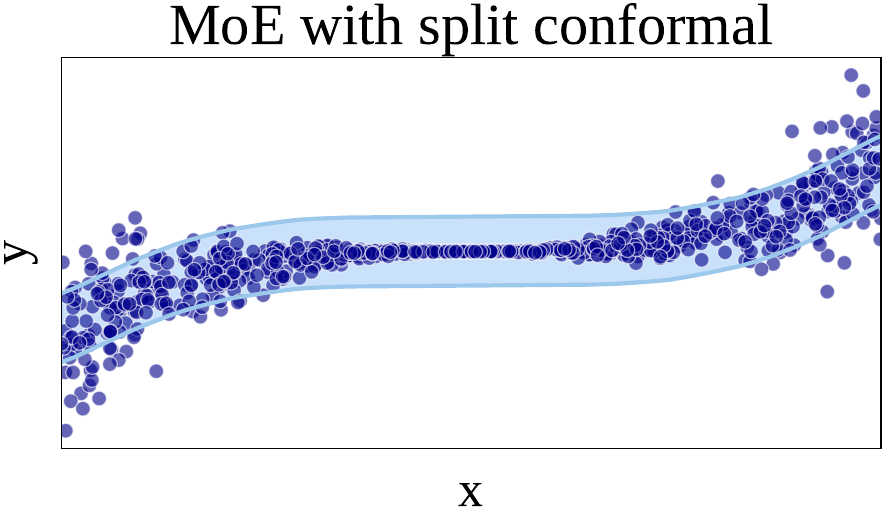}
    \hfill
    \includegraphics[width=0.49\textwidth]{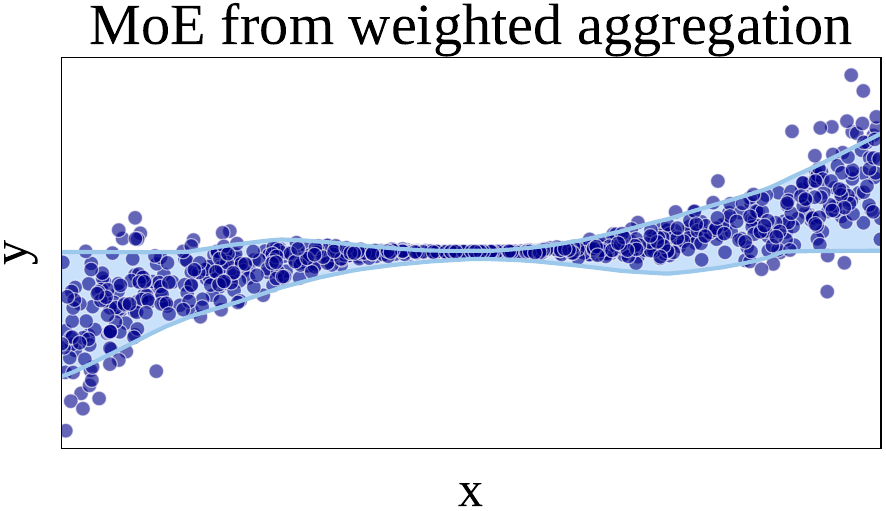}
\end{minipage}

\vspace{0pt} 

\caption{
\textbf{Left:} Network diagram for MoE. For traditional MoE, the aggregation module takes a weighted sum of the outputs from each expert. To learn weight-dependent prediction sets, we instead propose to combine the prediction sets of each expert by weighted p-value.
\textbf{Right, top row:} Comparison of split conformal prediction sets with those learned from weighted aggregation. Weighted aggregation allows overall coverage to follow the coverage of the dominant expert, rather than remain purely marginal.
\textbf{Right, bottom row:} Another comparison of split conformal with weighted aggregation, with the latter showing local coverage with smooth transitions.
\vspace{-8pt} 
}
\label{fig:moe_conditional_comparison}
\end{figure*}

The routing network's ability to learn data-dependent weights for each expert makes MoE a natural setting for applying our method of aggregating prediction sets by weighted p-values. We collect the routing weights into the simplex-valued vector $W(x) := (W_1(x), \dots, W_K(x))$; then, for a new input $X_{n+1}$, the routing network outputs the weight vector $\mathbf{W}^{(n+1)} := W(X_{n+1})$, whose $k$th component we denote $W^{(n+1)}_k$. Let $\ph_k$ denote the p-value function of the $k$th expert. The weighted aggregate p-value function for label $y$ is
\begin{equation*}
p_{\text{all}}(X_{n+1}, y; \mathbf{W}^{(n+1)}) = \sum_k W_k^{(n+1)} \ph_k(X_{n+1}, y).
\end{equation*}
At significance level $\alpha$, we form the MoE prediction set by thresholding the corrected p-variable $\widehat m^* p_{\text{all}}$, as established in Proposition~\ref{prop:coverage_avg_dataw_empirical}:
\begin{equation*}
\Ch_{\alpha}^{\text{MoE}}(X_{n+1}) = \{y \in \mathcal{Y} : \widehat{m}^* p_{\text{all}}(X_{n+1}, y; \mathbf{W}^{(n+1)}) > \alpha\}.
\end{equation*}
Intuitively, the routing weights adapt the contribution of each expert’s p-value to the input so that experts with higher predictive relevance for $X_{n+1}$ have greater influence. This pairs naturally with weighted p-value aggregation for two reasons: (1) averaging p-values retains graded evidence from each expert, so that the weight-aggregated p-value faithfully reflects expert disagreement; (2) although data-dependent weights generally break p-value validity, the p-value formulation admits a scaling correction \eqref{eq:m_correction_factor} that restores finite-sample validity, even though the routing weights are learned from the data. We expand on both these points in \S\ref{app:other_baselines_benefits}, and we illustrate the resulting gains in local validity in Figure~\ref{fig:moe_conditional_comparison} and in the following experiments.

\paragraph{Baselines}
We refer to our proposed method of aggregating expert p-value functions with learned weights as \textit{weighted aggregation}. As our main baseline, we compare against split conformal with the full MoE predictor, and we refer to this simply as \textit{split conformal} for brevity.

To evaluate local validity, we compare against conformal quantile regression (CQR) \citep{romano2019conformalized}, a widely used locally adaptive method. A strength of our framework is that it complements, rather than competes with, other adaptive methods, allowing weighted aggregation to be layered on top of CQR. Accordingly, we also evaluate a hybrid method that combines weighted aggregation with CQR. We assess performance using marginal coverage, worst-slice (WS) coverage \citep{romano2020classification}, and prediction set size.

We consider two practical variants of weighted aggregation:
\begin{itemize}[leftmargin=15pt,topsep=0pt,itemsep=0mm]
    \item \textit{WA targeted $(0, \alpha']$}: coverage is guaranteed for all $\alpha \in (0, \alpha')$ using the $m^\dag$ correction \eqref{eq:m_correction_factor_tighter}.
    \item \textit{WA precise $\alpha'$}: coverage is guaranteed for $\alpha = \alpha'$ only, using the $m^\ddag$ correction \eqref{eq:m_correction_factor_exact}.
\end{itemize}

\begin{figure*}[t]
    \centering
    \includegraphics[width=\textwidth]{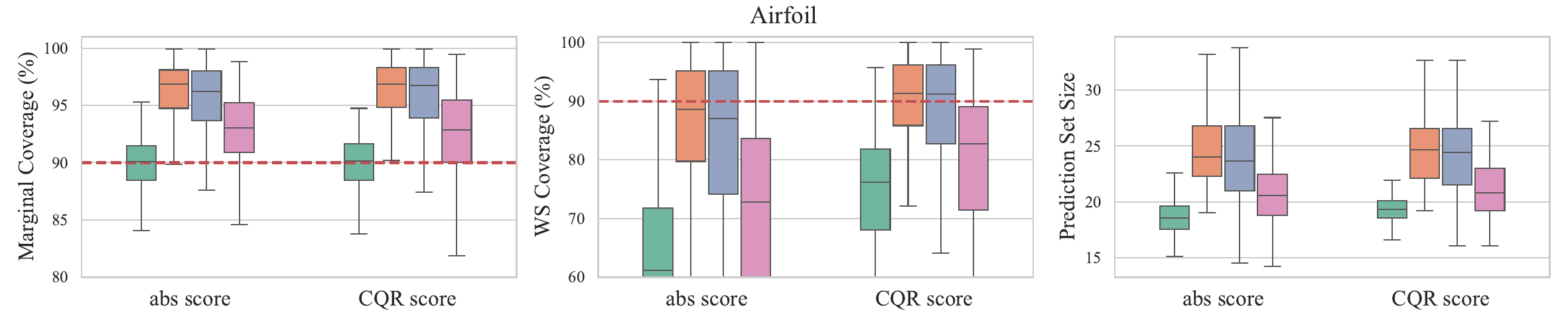} \\ \vspace{-1pt}
    \includegraphics[width=\textwidth]{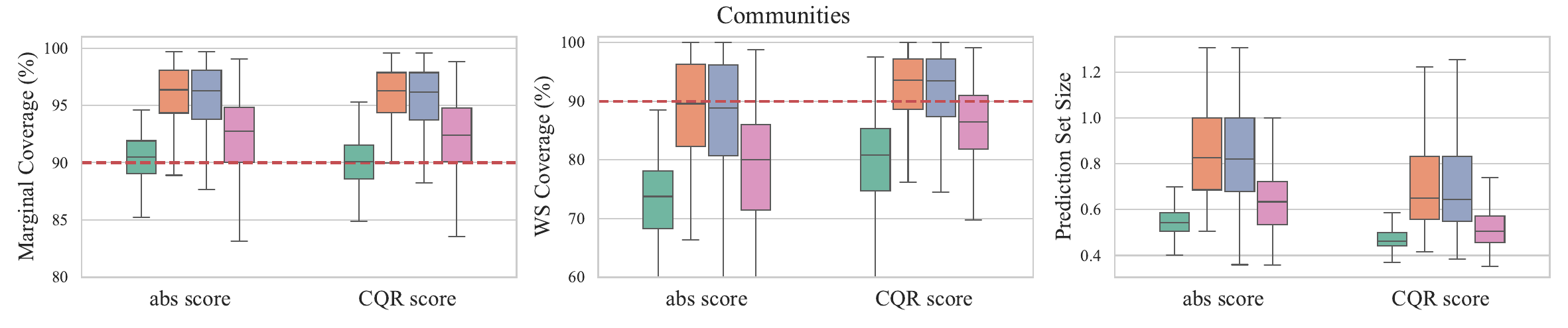} \\ \vspace{-1pt}
    \includegraphics[width=\textwidth]{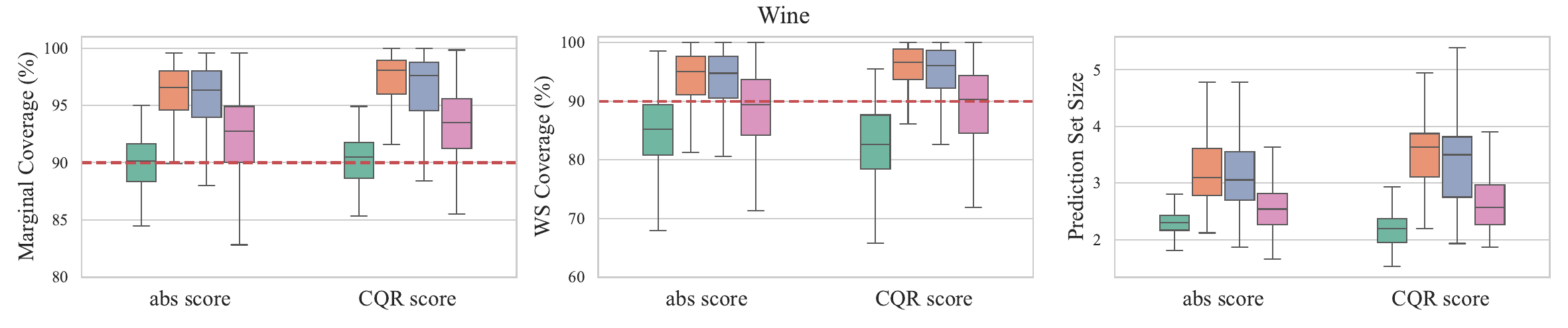} \\ \vspace{-1pt}
    \includegraphics[width=\textwidth]{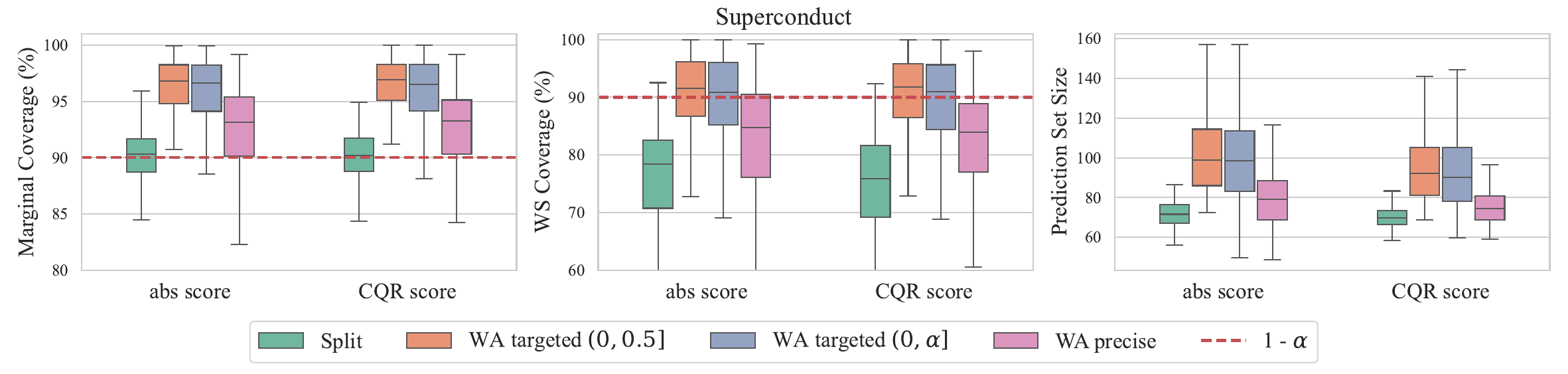} \\ \vspace{-1pt}
    
    \caption{Local validity experiments comparing split conformal to weighted aggregation using absolute residual scores and CQR scores. Each row corresponds to a dataset, with plots for marginal coverage, WS coverage, and prediction set size from left to right. The weighted aggregation methods consistently improves WS coverage across datasets, in most cases meeting target coverage on the WS slab. Split conformal is less conservative marginally, but undercovers on the WS slab; using CQR scores is not enough to cover the gap.
    \vspace{-10pt} 
    }
    \label{fig:uci_regression_plots}
\end{figure*}

\paragraph{Experiment overview}
For clarity of exposition, we present our main experimental results on the local validity of our method for real-world regression tasks in Figure \ref{fig:uci_regression_plots}, and defer additional ablation studies and further analyses on local validity to the Appendix.

Figure \ref{fig:uci_regression_plots} summarizes our main regression results, with CQR as our main adaptive baseline. We also evaluate our method against additional data-adaptive baselines in \S\ref{app:exp_real_conservative_comparison} and \S\ref{app:other_baselines_adaptive}, and we present results on classification tasks in \S\ref{app:exp_real_class}.

Our primary metric for evaluating local validity is WS coverage. To complement this, we also examine coverage disparities across demographic groups on the Communities dataset \citep{redmond2002data} in \S\ref{app:exp_real_communities}

In Appendix \ref{app:additional_exp}, we analyze how various factors impact coverage and prediction set size in a series of ablation studies on synthetic data. We investigate: the effects of merging set size $|S_{\text{merge}}|$ (\S \ref{app:synthetic_exp_smerge}), shared feature information (\S \ref{app:synthetic_feature_assignment}), and the different weighted aggregation variants (\S \ref{app:synthetic_guarantees}).

The following experiments are conducted with an MoE where the routing network and experts are linear models for simplicity. All results are averaged over 200 trials.

\paragraph{Main regression results}
\label{sec:exp_real_reg}
In Figure \ref{fig:uci_regression_plots}, we compare split conformal to weighted aggregation using both absolute residual scores and CQR scores. From the marginal coverage plots (left column), we see that split conformal consistently achieves coverage closest to the nominal $1-\alpha$ level, while the weighted aggregation methods tend to overcover. However, the WS coverage plots (middle column) expose a well-known limitation of split conformal: it does not maintain coverage across the data, systematically undercovering on the WS slab. CQR scores improve split conformal WS coverage on some datasets (Airfoil and Communities), but they are still insufficient to close the coverage gap.

Weighted aggregation significantly improves WS coverage, coming close to or meeting the nominal level on all datasets. In addition, the gap between marginal and WS coverage is notably smaller for weighted aggregation than for split conformal, indicating that weighted aggregation provides more uniform coverage, even over the challenging regions of the data. These results suggest that weighted aggregation is preferable in applications where local validity is a priority.

The prediction set size plots (right column) illustrate the standard trade-off between coverage and efficiency, where higher-coverage methods like WA targeted tend to produce larger prediction sets than lower-coverage methods like split conformal. According to these results, we suggest WA targeted if maintaining coverage in challenging regions is the primary concern. If both efficiency and local validity are important, WA precise provides a reasonable middle ground---offering improved WS coverage over split conformal, while also maintaining more compact prediction set size.

\section{Conclusion}
The asymmetric (weighted) aggregation of prediction sets is a flexible generalization of standard symmetric aggregation. Intuitively, coverage in this setting should vary with the distribution of weights. The results of \citet{vovk2020combining} allow us to formalize this intuition for conformal prediction, so that for data-independent weights based on expert priors, we obtain improved guarantees when the weights are sufficiently asymmetric. We extend this framework to data-dependent weights (e.g. weights learned from data), enabling adaptive coverage that reflects the observed input. Experiments on WS coverage and demographic subgroups confirm the practical benefits of this extension, showing that weighted aggregation yields more reliable coverage in challenging settings.

\section*{Acknowledgements}
We thank Josh Gleason and Uthsav Chitra for their thoughtful feedback, and Josh Gleason in particular for many insightful discussions throughout this work. GW and RC are partially supported by a ONR MURI grant N00014-20-1-2781. DP is partially supported by an Amazon AI2AI fellowship. DP, SS, and AL are partially supported by the Gordon and Betty Moore Foundation grant \#12128. AL is also supported by an Amazon Research Award. 

\bibliographystyle{unsrtnat}
\bibliography{references}
\appendix
\section{Proofs and additional theoretical details} \label{app:proofs}

\subsection{Proof of Proposition \ref{prop:coverage_avg_constantw}} \label{app:proof1}
Following the notation of \citet{vovk2020combining}, we define the merging function
\begin{equation*}
M_{1,\mathbf{v}}(p_1, \dots, p_K) = (v_1 p_1 + \dots + v_K p_K) \quad\text{where } \sum_{k=1}^K v_k = 1.
\end{equation*}
Recall that we use p-variables to represent our p-value functions applied to data; that is, $P_k = \ph_k(X,Y)$. Then $\bar{P} = M_{1,\mathbf{v}}(P_1, \dots, P_K)$. By Proposition 9 of \citet{vovk2020combining}, $A_\mathbf{v} M_{1, \mathbf{v}}$ is a precise merging function, where $A_\mathbf{v} = \min\left\{\frac{1}{v},2\right\}$ and $v = \max\{v_1, ..., v_K\}$. Thus, $A_\mathbf{v} \bar{P}$ is a p-variable, and $\mathbb{P}\left\{A_\mathbf{v} \bar{P} \le \alpha\right\} \le \alpha$, or
\begin{equation*}
\mathbb{P}\left\{\bar{P} \le \alpha\right\} \le A_\mathbf{v} \alpha = \min \left\{ \frac{1}{v}, 2 \right\} \alpha.
\end{equation*}
Under exchangeability, the prediction set
\begin{equation*}
\Ch_{\alpha}^{\text{avg}}(X_{n+1}) = \{y \in \mathcal{Y} : \bar{p}(X_{n+1}, y) > \alpha\}
\end{equation*}
constructed from $\bar{p}$ has the coverage guarantee
\begin{equation*}
\mathbb{P} \left\{ Y_{n+1} \in \Ch^{\text{avg}}_{\alpha}(X_{n+1}) \right\} \ge 1 - \min \left\{ \frac{1}{v}, 2 \right\} \alpha.
\end{equation*}

\subsection{Proof of Proposition \ref{prop:coverage_avg_dataw}} \label{app:proof2}
We define $m^*$ to be the smallest positive value such that $m^* P_{\text{all}}$ is a p-variable, i.e.\
\begin{equation} \label{eq:m_pvalue}
\mathbb{P}\left\{ m^* P_{\text{all}} \le \alpha \right\} \le \alpha,
\end{equation}
or, equivalently,
\begin{equation*}
\mathbb{P}\left\{ m^* P_{\text{all}} > \alpha \right\} \ge 1 - \alpha.
\end{equation*}
Recall that $P_{\text{all}} := p_{\text{all}}(X,Y,\mathbf{W})$. Then under exchangeability, the prediction set
\begin{equation*}
\Ch^{\text{scaled}}_{\alpha}(X_{n+1}) = \left\{ y \in \mathcal{Y} : m^* p_{\text{all}}(X_{n+1}, y; \mathbf{W}^{(n+1)}) > \alpha \right\}
\end{equation*}
has coverage
\begin{equation*}
\mathbb{P} \left\{ Y_{n+1} \in \Ch^{\text{scaled}}_{\alpha}(X_{n+1}) \right\} \ge 1 - \alpha.
\end{equation*}

Alternatively, we can rearrange \eqref{eq:m_pvalue} to be
\begin{equation*}
\mathbb{P}\left\{ P_{\text{all}} \le \frac{\alpha}{m^*} \right\} \le \alpha;
\end{equation*}
if we define $\alpha' = \alpha / m^*$, this becomes
\begin{equation*}
\mathbb{P}\left\{ P_{\text{all}} \le \alpha' \right\} \le m^* \alpha'
\end{equation*}
or
\begin{equation*}
\mathbb{P}\left\{ P_{\text{all}} > \alpha' \right\} \ge 1 - m^* \alpha'.
\end{equation*}
Under exchangeability, the prediction set
\begin{equation*}
\Ch^{\text{unscaled}}_{\alpha}(X_{n+1}) = \left\{ y \in \mathcal{Y} : p_{\text{all}}(X_{n+1}, y; \mathbf{W}^{(n+1)}) > \alpha \right\}
\end{equation*}
satisfies
\begin{equation*}
\mathbb{P} \left\{ Y_{n+1} \in \Ch^{\text{unscaled}}_{\alpha}(X_{n+1}) \right\} \ge 1 - m^* \alpha.
\end{equation*}
Note that other transformations of $P_{\text{all}}$ can also yield valid p-variables while preserving the relative proportion of weights. For example, if $P_{\text{all}}$ is shifted rather than scaled, then the condition
\begin{equation*}
\mathbb{P}\left\{ m' + P_{\text{all}} \le \alpha \right\} \le \alpha
\end{equation*}
also leads to coverage of at least $1 - \alpha$ for the scaled set, and at least $1 - (m' + \alpha)$ for the unscaled set. We focus on the correction factor $m^*$ because it aligns with the framework of \citet{vovk2020combining} and works well in practice.

\subsection{Proof of Proposition \ref{prop:coverage_avg_dataw_empirical}} \label{app:proof3}
Let $\mathcal{G}$ be the event that the worst-case distance between the true CDF $F_{P_{\text{all}}}$ and the empirical CDF $\widehat{F}_{P_{\text{all}}}$ is bounded by some maximum allowable deviation; that is,
\begin{equation*}
    \mathcal{G} = \left\{ \sup_{x \in \mathbb{R}} \left| \widehat{F}_{P_{\text{all}}}(x) - F_{P_{\text{all}}}(x) \right| \leq \varepsilon \right\} \quad \text{where } \varepsilon = \sqrt{\frac{\log(2/\delta)}{2 |S_{\text{merge}}|}}
\end{equation*}
for some user-chosen failure probability $\delta \in (0,1)$. By the Dvoretzky–Kiefer–Wolfowitz (DKW) inequality, 
\begin{equation*}
    \mathbb{P}(\mathcal{G}) \ge 1 - \delta.
\end{equation*}

Now, suppose we have a fixed merging set $S_{\text{merge}}$ such that $\mathcal{G}$ holds. By definition of a CDF, 
\begin{equation} \label{eq:prob_to_cdf}
    \mathbb{P} \left\{\widehat{m}^* P_{\text{all}} \leq \alpha \mid S_{\text{merge}} \right\} = F_{P_{\text{all}}} \left(\frac{\alpha}{\widehat{m}^*} \right),
\end{equation}
and under the same fixed $S_{\text{merge}}$,
\begin{equation*}
    F_{P_{\text{all}}}(x) \le \widehat{F}_{P_{\text{all}}}(x) + \varepsilon.
\end{equation*}
We take $x = \alpha / \widehat{m}^*$; this allows us to express the bound as
\begin{equation} \label{eq:cdf_oneside_dkw_bound}
    F_{P_{\text{all}}}\left(\frac{\alpha}{\widehat{m}^*}\right) \le \widehat{F}_{P_{\text{all}}}\left(\frac{\alpha}{\widehat{m}^*}\right) + \varepsilon.
\end{equation}
By construction of $\widehat{m}^*$ \eqref{eq:empirical_m_correction_factor}, we know that $\widehat{F}_{P_{\text{all}}}(x)/x \le \widehat{m}^*$ for all $x$. Substituting again $x = \alpha/\widehat{m}^*$, we have
\begin{equation} \label{eq:empirical_cdf_bound}
    \widehat{F}_{P_{\text{all}}}\left(\frac{\alpha}{\widehat{m}^*}\right) \le \widehat{m}^* \frac{\alpha}{\widehat{m}^*} = \alpha.
\end{equation}
Combining \eqref{eq:cdf_oneside_dkw_bound} and \eqref{eq:empirical_cdf_bound} gives 
\begin{equation*}
    F_{P_{\text{all}}} \left(\frac{\alpha}{\widehat{m}^*} \right) \leq \alpha + \varepsilon.
\end{equation*}
Applying this to \eqref{eq:prob_to_cdf} gives
\begin{equation}\label{eq:pvariable_Smerge}
    \mathbb{P} \left\{\widehat{m}^* P_{\text{all}} \leq \alpha \mid S_{\text{merge}} \right\} \leq \alpha + \varepsilon.
\end{equation}
Under exchangeability, the prediction set
\begin{equation*}
\Ch^{\text{scaled}}_{\alpha}(X_{n+1}) = \left\{ y \in \mathcal{Y} : \widehat{m}^* p_{\text{all}}(X_{n+1}, y; \mathbf{W}^{(n+1)}) > \alpha \right\}
\end{equation*}
has a conditional \textit{miscoverage} guarantee of
\begin{equation*}
    \mathbb{P} \left\{ Y_{n+1} \notin \Ch^{\text{scaled}}_\alpha(X_{n+1}) \mid S_{\text{merge}} \right\} \leq \alpha + \varepsilon.
\end{equation*}
Let us denote the miscoverage indicator as
\begin{equation*}
    \text{MC} = \mathds{1}\left\{ Y_{n+1} \notin \Ch^{\text{scaled}}_\alpha(X_{n+1}) \right\}.
\end{equation*}
The indicator random variable allows us to easily switch between probability and expectation in order to marginalize over $S_{\text{merge}}$:
\begin{equation*}
\begin{aligned}
\mathbb{P} \left\{ \text{MC}=1 \right\} &= \mathbb{E}\left[ \text{MC} \right] \\
&= \mathbb{E}\left[ \mathbb{E}\left[ \text{MC} \mid S_{\text{merge}} \right] \right] \\
&= \mathbb{E}\left[\mathbb{P}\left\{ \text{MC}=1 \mid S_{\text{merge}} \right\}\right] \\
&\leq \left(\alpha + \varepsilon\right) \mathbb{P}(\mathcal{G}) + 1 \cdot \mathbb{P}(\mathcal{G}^c) \\
&\leq \alpha + \varepsilon + \delta.
\end{aligned}
\end{equation*}
This gives the coverage guarantee
\begin{equation*}
    \mathbb{P} \left\{ Y_{n+1} \in \Ch^{\text{scaled}}_\alpha(X_{n+1}) \right\} \geq 1 - \left(\alpha + \varepsilon + \delta \right).
\end{equation*}
To derive the guarantee for the \textit{unscaled} prediction set, we apply a change of variable to \eqref{eq:pvariable_Smerge} to get
\begin{equation*}
    \mathbb{P} \left\{ P_{\text{all}} \leq \alpha' \mid S_{\text{merge}} \right\} \leq \widehat{m}^* \alpha' + \varepsilon.
\end{equation*}
The remainder of the proof follows the same conditioning and marginalization argument, with the substitution of $\widehat{m}^*\alpha$ for $\alpha$ carried through. To be explicit, the prediction set
\begin{equation*}
\Ch^{\text{unscaled}}_{\alpha}(X_{n+1}) = \left\{ y \in \mathcal{Y} : p_{\text{all}}(X_{n+1}, y; \mathbf{W}^{(n+1)}) > \alpha \right\}
\end{equation*}
has a conditional miscoverage guarantee of
\begin{equation*}
    \mathbb{P} \left\{ Y_{n+1} \notin \Ch^{\text{unscaled}}_\alpha(X_{n+1}) \mid S_{\text{merge}} \right\} \leq \widehat{m}^* \alpha + \varepsilon.
\end{equation*}
Then miscoverage indicator $\text{MC}' = \mathds{1}\left\{ Y_{n+1} \notin \Ch^{\text{unscaled}}_\alpha(X_{n+1}) \right\}$ can help us marginalize over $S_{\text{merge}}$:
\begin{equation*}
\begin{aligned}
\mathbb{P} \left\{ \text{MC}'=1 \right\} &= \mathbb{E}\left[\mathbb{P}\left\{ \text{MC}'=1 \mid S_{\text{merge}} \right\}\right] \\
&\leq \left(\alpha \mathbb{E}[\widehat{m}^*] + \varepsilon\right) \mathbb{P}(\mathcal{G}) + 1 \cdot \mathbb{P}(\mathcal{G}^c) \\
&\leq \alpha \mathbb{E}[\widehat{m}^*] + \varepsilon + \delta.
\end{aligned}
\end{equation*}
This gives the guarantee
\begin{equation*}
    \mathbb{P} \left\{ Y_{n+1} \in \Ch^{\text{unscaled}}_\alpha(X_{n+1}) \right\} \geq 1 - \left(\alpha \mathbb{E}[\widehat{m}^*] + \varepsilon + \delta \right).
\end{equation*}
Note that the same argument applies directly if we replace the empirical CDF $\widehat{F}_{P_{\text{all}}}$ by the conservative version $\widehat{F}_{P_{\text{all}}}^{\text{cons}}$ \eqref{eq:cons_empirical_cdf} because the two differ by at most $(|S_{\text{merge}}| + 1)^{-1}$, a deterministic offset that can simply be absorbed into the DKW tolerance by replacing $\varepsilon$ with $\varepsilon + (|S_{\text{merge}}| + 1)^{-1}$.

\subsection{\texorpdfstring{More detailed derivation of $m^*$}{More detailed derivation of m*}} \label{app:derivation_mstar}
For some distribution function $F_{P_{\text{all}}}$, we define
\begin{equation*}
    m^* = \inf\left\{m>0 : F_{P_{\text{all}}}(\alpha/m)\le \alpha \text{ for all }\alpha\in(0,1)\right\}, \qquad c = \sup_{\delta>0}\frac{F_{P_{\text{all}}}(\delta)}{\delta}.
\end{equation*}
We aim to prove that $m^*=c$ to establish the equivalence in \eqref{eq:m_correction_factor}. To aid our proof, we define the feasible set
\begin{equation*}
    S = \left\{m>0 : F_{P_{\text{all}}}(\alpha/m)\le \alpha \text{ for all }\alpha\in(0,1)\right\},
\end{equation*}
where $m^* = \inf(S)$.

We begin by showing that $m^* \leq c$. To this end, consider any $m>c$. By definition of $c$,
\begin{equation*}
    F_{P_{\text{all}}}(\delta)\le c\,\delta < m\,\delta \quad \forall \delta > 0.
\end{equation*}  

Pick $\delta=\alpha/m$ for any $\alpha \in (0,1)$ so that
$\delta \in (0, 1/m)$. Substituting yields
\begin{equation*}
    F_{P_{\text{all}}}(\alpha/m) < m (\alpha / m) = \alpha.
\end{equation*}
This shows that every $m>c$ satisfies the feasibility condition, so every $m>c$ is in $S$; it follows that $S \supset (c, \infty)$. Then, as the infimum of $S$, $m^*\le c$.

To show that $m^* \geq c$, consider any feasible $m > 0$, i.e.\ assume
\begin{equation*}
    F_{P_{\text{all}}}(\alpha/m)\le\alpha \quad \forall \alpha\in(0,1).
\end{equation*}

For any $\delta>0$, there are two cases.
\begin{enumerate}
\item If $\delta \ge 1/m$, since $F_{P_{\text{all}}}(\delta)\le 1$, 
\begin{equation*}
    F_{P_{\text{all}}}(\delta)/\delta\le 1/\delta\le m.
\end{equation*}

\item If $\delta \in (0, 1/m)$, let us select $\alpha = m\delta \in (0,1)$. By feasibility of $m$,
\begin{equation*}
F_{P_{\text{all}}}(\delta) = F_{P_{\text{all}}}(\alpha/m)\le\alpha = m\delta,
\end{equation*}
or
\begin{equation*}
F_{P_{\text{all}}}(\delta)/\delta\le m.
\end{equation*}
\end{enumerate}
The two cases above establish $m \ge c$ for every $m \in S$, or that $S \subset (c, \infty)$. Thus, $c$ is a lower bound of $S$ and $m^* \ge c$.

\subsection{\texorpdfstring{Computing $\widehat{m}^*$ from an empirical CDF}{Computing mhat* from an empirical CDF}} \label{app:computing_mhatstar}
For random variable $P_{\text{all}} = p_{\text{all}}(X, Y; \mathbf{W})$, 
a \text{merging correction factor} $\lambda$ is a positive scalar that ensures that a $\lambda P_{\text{all}}$ is a valid p-value. That is,
\begin{equation*}
    \lambda \in \{ m > 0 : \mathbb{P}\{m P_{\text{all}} \leq \alpha \} \leq \alpha ~ \forall \alpha \in (0, 1) \} = \{ m > 0 : F_{P_{\text{all}}}\left(\alpha/m\right) \leq \alpha ~ \forall \alpha \in (0, 1) \}.
\end{equation*}
Scaling by a merging correction factor gives us the guarantee
\begin{equation*}
    \mathbb{P}\{ P_{\text{all}} > \alpha \} = 1 - \lambda \alpha.
\end{equation*}
To achieve the tightest guarantee (with application to all $\alpha \in (0,1)$), we define the minimal merging correction factor $m^*$ to be
\begin{equation} \label{eq:scale_sup}
    m^{*} = \inf\{m > 0 : F_{P_{\text{all}}}\left(\alpha/m\right) \leq \alpha ~ \forall \alpha \in (0, 1) \} = \sup_{\delta > 0}{\frac{F_{P_{\text{all}}}(\delta)}{\delta}}.
\end{equation}
(Equation \eqref{eq:scale_sup} is a restatement of \eqref{eq:m_correction_factor} from the main body; we replicate it here for easy reference.)

In practice, we compute the minimal merging factor $\widehat{m}^*$ by first constructing the empirical CDF of $P_{\text{all}}$ over the merging set
\begin{equation*}
\widehat{F}_{P_{\text{all}}}(\alpha) = \frac{\sum_{i \in S_\text{merge}} \mathds{1}\left\{ p_{\text{all}}(X_i, Y_i; \mathbf{W}^{(i)}) \leq \alpha \right\}}{|S_\text{merge}|},
\end{equation*}
or its conservative version $\widehat{F}^{\text{cons}}_{P_{\text{all}}}$ \eqref{eq:cons_empirical_cdf}. For the sake of conciseness, let
\begin{equation*}
    \widehat{F}_i := \widehat{F}_{P_{\text{all}}}\left(p_{\text{all}}(X_i, Y_i; \mathbf{W}^{(i)})\right).
\end{equation*}
Since $\widehat{F}_{P_{\text{all}}}$ is a right-continuous step function that only jumps at the observed values $\{p_{\text{all}}(X_i, Y_i; \mathbf{W}^{(i)}) : i \in S_{\text{merge}}\}$, the supremum in \eqref{eq:scale_sup} is attained at one of these points. In fact, for any $\delta$ not equal to one of these values, shifting $\delta$ slightly to the right (toward the next jump) increases the denominator without changing the numerator, decreasing the ratio. For this reason, it suffices to take the maximum over the observed p-values, yielding the empirical merging factor
\begin{equation*}
\widehat{m}^* = \max_{\substack{i \in S_{\text{merge}} \\[1pt] p_{\text{all}}(X_i, Y_i; \mathbf{W}^{(i)}) > 0}} \frac{\widehat{F}_i}{p_{\text{all}}(X_i, Y_i; \mathbf{W}^{(i)})}.
\end{equation*}
In some cases, we are only interested in coverage above a certain significance level $\alpha$ as in \eqref{eq:m_correction_factor_tighter}. Then we only need to find the first point on the empirical CDF that surpasses $\alpha$, or
\begin{equation*}
    \bar{\alpha} = \operatornamewithlimits{min}_{\substack{i \in S_{\text{merge}} \\ \widehat{F}_i \geq \alpha}} \widehat{F}_i,
\end{equation*}
and then compute the scaling factor to be
\begin{equation*}
\widehat{m}^\dag = \max_{\substack{i \in S_{\text{merge}} \\[1pt] p_{\text{all}}(X_i, Y_i; \mathbf{W}^{(i)}) > 0 \\ \widehat{F}_i \leq \bar{\alpha}}} \frac{\widehat{F}_i}{p_{\text{all}}(X_i, Y_i; \mathbf{W}^{(i)})}.
\end{equation*}

\newpage
\section{Additional experiments} \label{app:additional_exp}

\subsection{Synthetic data} \label{app:exp_synthetic}
We generate a simple homoskedastic dataset to simulate a regression task, where each input is a 16-dimensional Gaussian with random parameters, and the label is the sum of the different dimensions with additive noise (see \S \ref{app:exp_details_synthetic} for more detail). We use this dataset to investigate how various factors impact the coverage and interval width of our prediction sets.

\subsubsection{\texorpdfstring{Coverage and size of prediction sets improve with larger $|S_{\text{merge}}|$}{Coverage and size of prediction sets improve with larger Smerge}} \label{app:synthetic_exp_smerge}
The merging set $S_{\text{merge}}$ allows us to construct an empirical CDF for $P_{\text{all}}$. To improve stability in finite samples, we apply a conservative correction to the typical formula for the empirical CDF, and use this conservative CDF \eqref{eq:cons_empirical_cdf} to compute the correction factor $\widehat{m}^*$ \eqref{eq:empirical_m_correction_factor}. Our conservative CDF tends to be overly conservative when $|S_{\text{merge}}|$ is small, leading to overcoverage. However, as $|S_{\text{merge}}|$ grows and the empirical CDF approaches the true CDF, coverage becomes closer to nominal and the prediction sets become less conservative.

\begin{figure}[ht]
    \centering
    \includegraphics[width=0.4\textwidth]{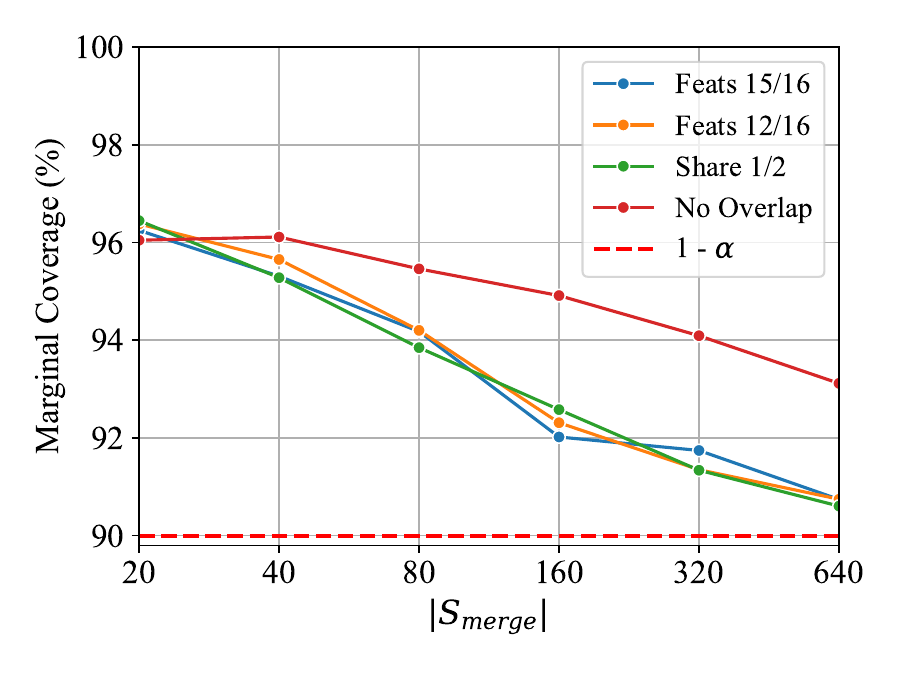}
    \hspace{10pt}
    \includegraphics[width=0.4\textwidth]{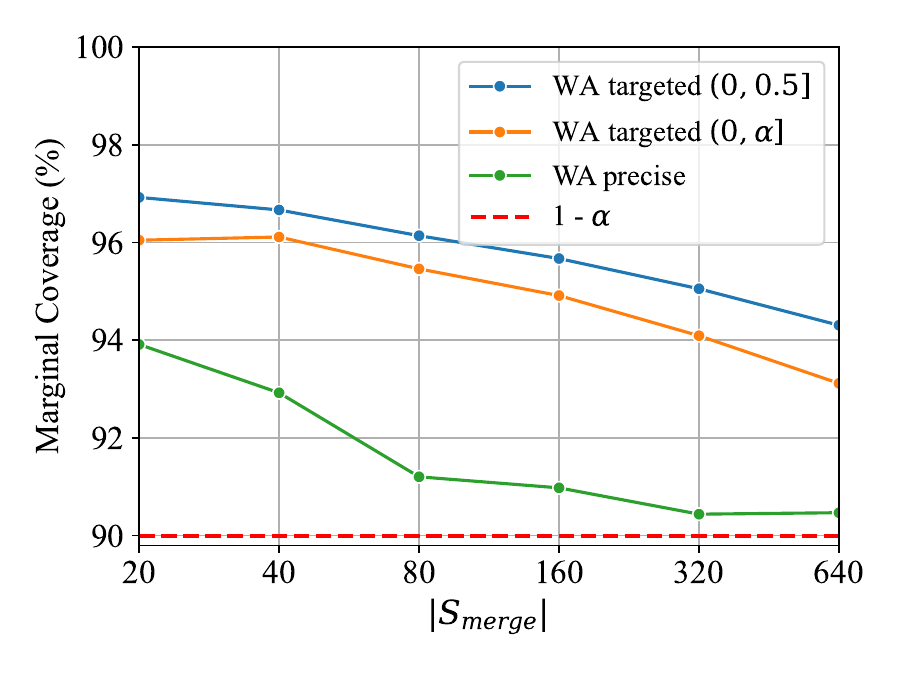}
    \vspace{-5pt}
    \caption{Mean coverage compared to the size of the merging set $S_{\text{merge}}$, with different feature assignments (left) and different weighted aggregation variants (right). Overall, we find that coverage improves as the merging set gets larger for all methods. We note that the merging set does not have to be prohibitively large to produce decent results: for example, most feature assignment methods overcover by only 2\% with fewer than 200 samples.}
    \label{fig:merging_set_size}
\end{figure}

Figure \ref{fig:merging_set_size} illustrates how the size of the merging set $|S_{\text{merge}}|$ impacts mean coverage. On the left plot, we use the $m^\dag$ correction \eqref{eq:m_correction_factor_tighter} for \textit{WA targeted} (that is, $(0,\alpha)$ aggregation), and we compare different ways of assigning features to experts. On the right plot, we use a non-overlapping feature assignment, and we compare the different variants of weighted aggregation. Our results show that larger $|S_{\text{merge}}|$ leads to tighter coverage, with 160 samples being sufficient to achieve $<3\%$ overcoverage for most feature assignment methods. Interestingly, configurations where experts have no overlapping features tend to overcover the most. We explore the effect of feature assignment in the next section.

\subsubsection{Feature information overlap leads to higher coverage and more efficient prediction sets} \label{app:synthetic_feature_assignment}
\begin{figure}[ht]
    \centering
    \includegraphics[width=0.4\textwidth]{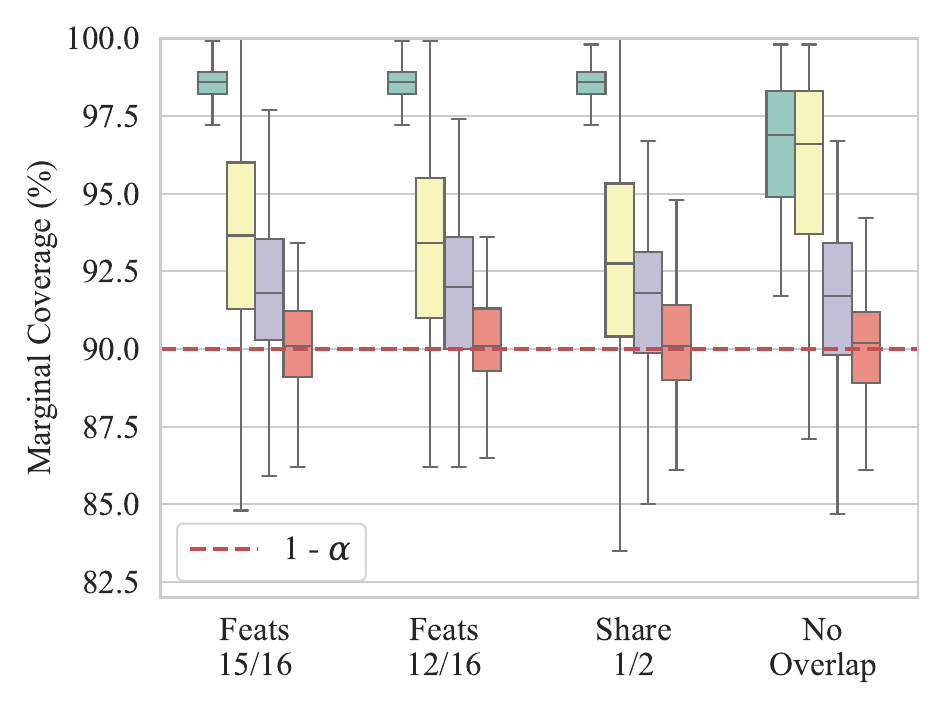}
    \hspace{10pt}
    \includegraphics[width=0.4\textwidth]{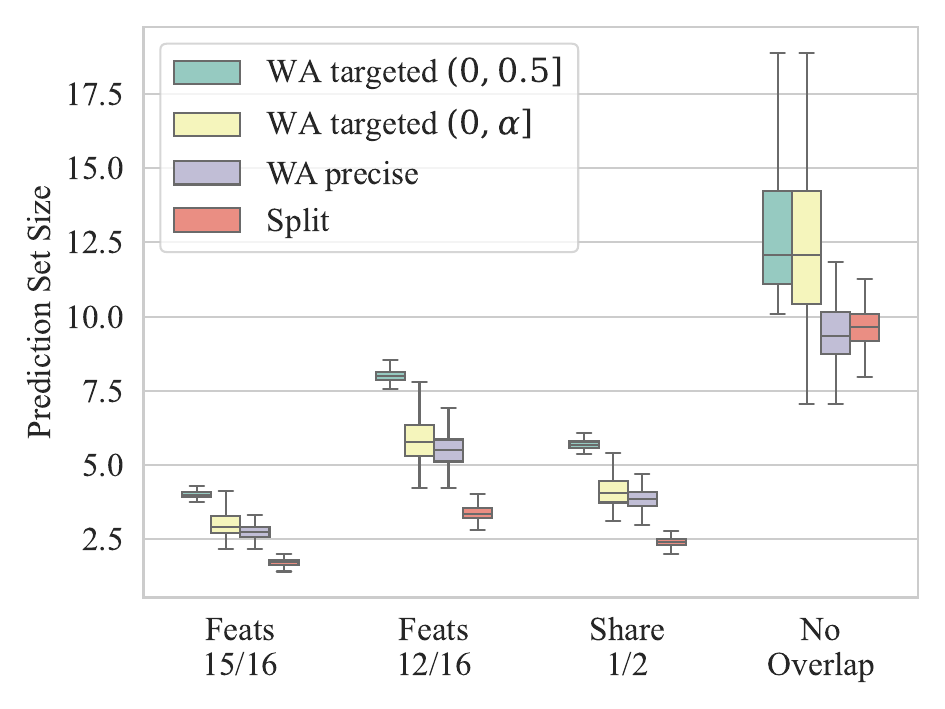}
    \vspace{-5pt}
    \caption{Coverage (left) and prediction set size (right) for different feature assignment methods and weighted aggregation methods. Our results indicate that sharing fewer features leads to tighter coverage, but sharing more features leads to more efficient (smaller) prediction sets. We also see that WA precise tends to have coverage that is closest to nominal and the most efficient prediction sets, albeit with a much looser guarantee.}
    \label{fig:featuredist_boxplots}
\end{figure}

To better understand how the allocation of features to experts affects the behavior of MoE weighted aggregation, we define four feature assignment methods (for our MoE of four experts):
\begin{itemize}
    \item \textit{Features 15/16}: each expert predicts from 15 of the 16 available features.
    \item \textit{Features 12/16}: each expert predicts from 12 of the 16 available features.
    \item \textit{Share 1/2}: all experts share 8 of the 16 features and partition the remaining 8 (2 features each).
    \item \textit{No Overlap}: the experts partition the 16 features (4 features each).
\end{itemize}
Figure \ref{fig:featuredist_boxplots} shows how different feature assignment methods affect coverage and prediction set size. Broadly, we observe that greater feature overlap leads to higher coverage (exceeding the nominal level) and more efficient prediction sets. In the MoE setting, this may be because feature sharing leads to more consistent estimations across experts, improving the reliability of the aggregated p-values and, in turn, reducing the size of the prediction sets. More generally, this may imply that the information redundancy introduced by feature overlap allows for better sample efficiency. This parallels findings in aggregation methods like cross-conformal and jackknife+, which also tend to produce smaller prediction sets than split conformal by reusing data. These results suggest that feature sharing is an important design consideration when aggregating prediction sets from multiple models.

\subsubsection{More general coverage guarantees result in more conservative prediction sets} \label{app:synthetic_guarantees}
We now compare the different variants of weighted aggregation described in \S \ref{sec:moe}, focusing on how the generality of the guarantee for each variant affects its empirical coverage.

\begin{figure}[ht]
    \centering
    \includegraphics[width=0.41\textwidth]{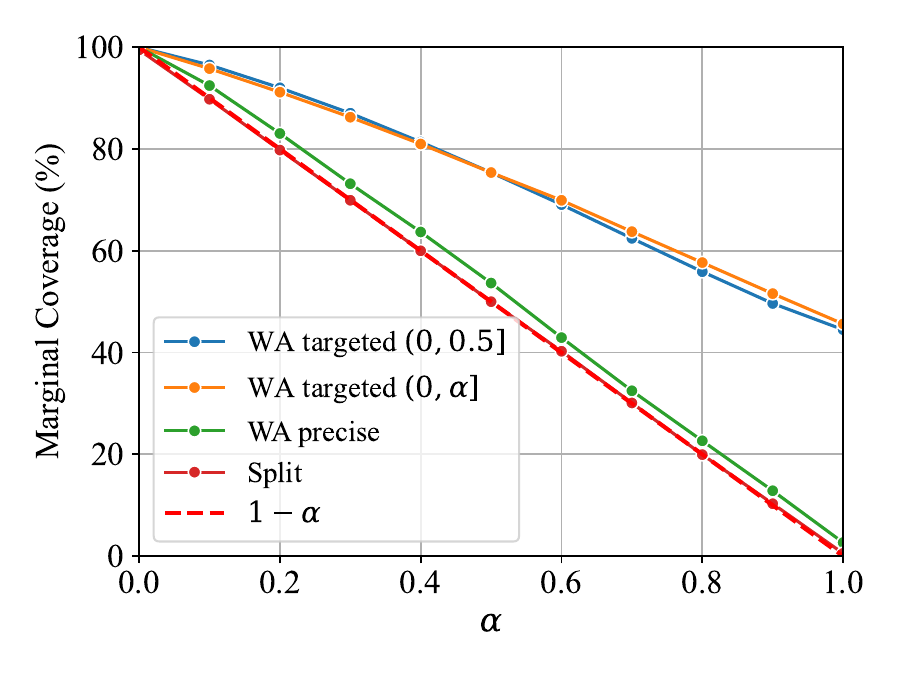}
    \vspace{-5pt}
    \caption{Mean coverage across significance levels $\alpha$ for the different weighted aggregation variants. WA precise achieves close to nominal coverage, while WA targeted tends to overcover (but offers more general guarantees).}
    \label{fig:alpha_v_coverage}
\end{figure}

Figure \ref{fig:featuredist_boxplots} compares the effects of different feature assignment methods and for each weighted aggregation variant (with $\alpha' = 0.1$), and Figure \ref{fig:alpha_v_coverage} shows how coverage varies across $\alpha$ for each variant.

Unsurprisingly, WA precise---which provides the narrowest guarantee, targeting a single $\alpha$---achieves coverage closest to the nominal level. In contrast, WA targeted offers more general guarantees over a range of $\alpha$ values, but tends to overcover, reflecting the conservativeness built into the method to accommodate worst-case behavior. We observe the same pattern on UCI data in Figure \ref{fig:uci_regression_plots}. These results illustrate the trade-off between the conservativeness of a method and the generality of its guarantee, and suggest that more targeted guarantees may be preferable when tighter coverage is important.

\subsection{Real data} \label{app:exp_real}

\subsubsection{Weighted aggregation improves local validity for classification} \label{app:exp_real_class}

\begin{figure}[ht]
    \centering
    \includegraphics[width=\textwidth]{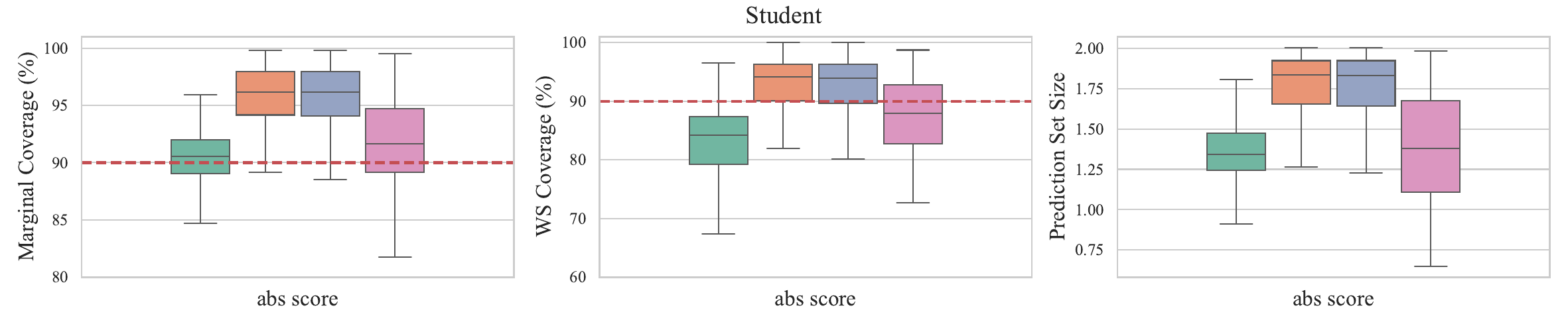} \\ \vspace{-2pt}
    \includegraphics[width=\textwidth]{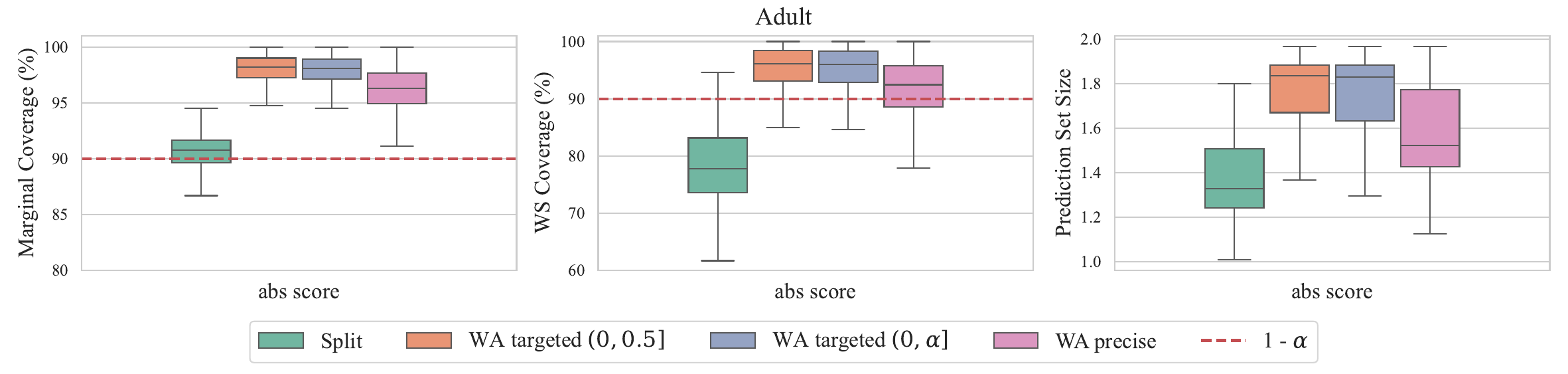}
    \caption{Local validity experiments comparing split conformal (green) to weighted aggregation (orange, purple, and pink) on the classification task. Like in Figure \ref{fig:uci_regression_plots}, each row corresponds to a dataset, with plots for marginal coverage, WS coverage, and prediction set size from left to right. We find that for classification as well as regression, WA improves WS coverage over split conformal, which undercovers in the WS region.}
    \label{fig:uci_classification_plots}
\end{figure}

The experiments in Figure \ref{fig:uci_classification_plots} mirror those in Figure \ref{fig:uci_regression_plots}, but for classification instead of regression. Like with the other set of experiments, we compare split conformal to weighted aggregation, and we see that split conformal achieves marginal coverage close to nominal but consistently undercovers on the WS slab. In contrast, weighted aggregation maintains much better WS coverage, suggesting it offers better local validity in the classification setting as well.

As before, we note that there is trade-off between coverage and efficiency. WA targeted is the best choice in terms of WS coverage, but WA precise balances the improved WS coverage of weighted aggregation methods with an efficiency that is closer to the split conformal. However, if marginal coverage and prediction set efficiency are the only priorities, then split conformal should be preferred.

\subsubsection{Local validity over demographic groups} \label{app:exp_real_communities}
\begin{figure}[ht]
    \centering
    \includegraphics[width=\textwidth]{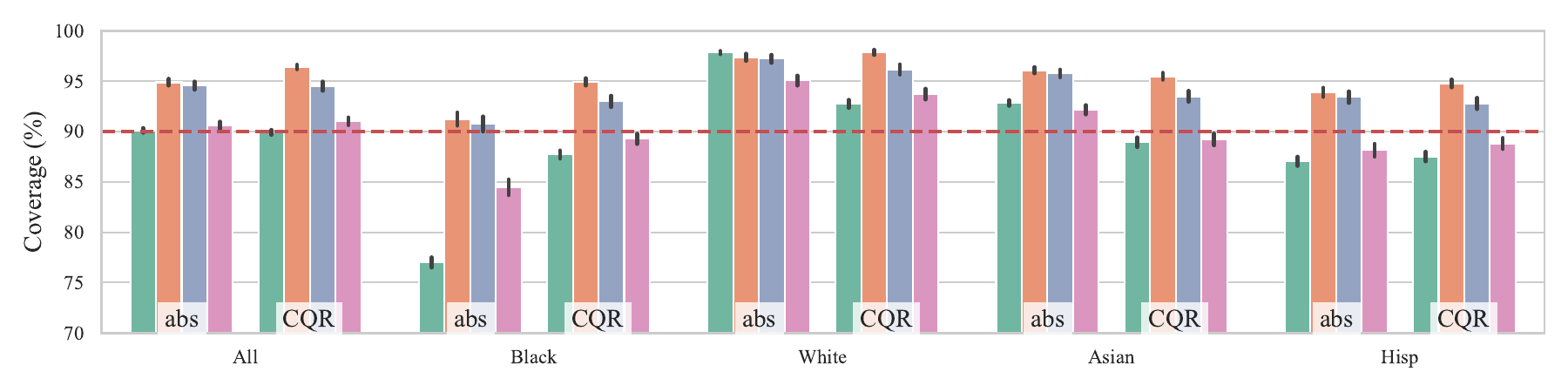} \\ \vspace{-2pt}
    \includegraphics[width=\textwidth]{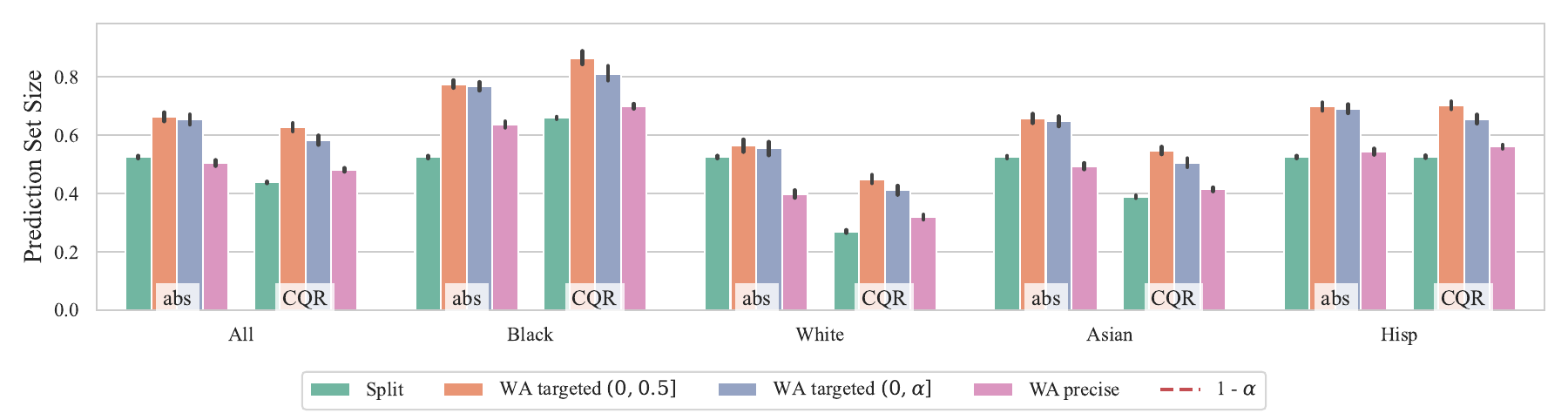} \\ \vspace{-2pt}
    \caption{Coverage (top) and prediction set size (bottom) for split conformal (green) and weighted aggregation variants (orange, purple, and pink) across subgroups with top 50th percentile racial representation. Split conformal has precise marginal coverage (``All''), but WA variants have more consistent coverage across subgroups, with WA targeted meeting coverage for all subgroups. Error bars represent 95\% confidence intervals.}
    \label{fig:app_exp_communities_50}
\end{figure}

\begin{figure}[ht]
    \centering
    \includegraphics[width=\textwidth]{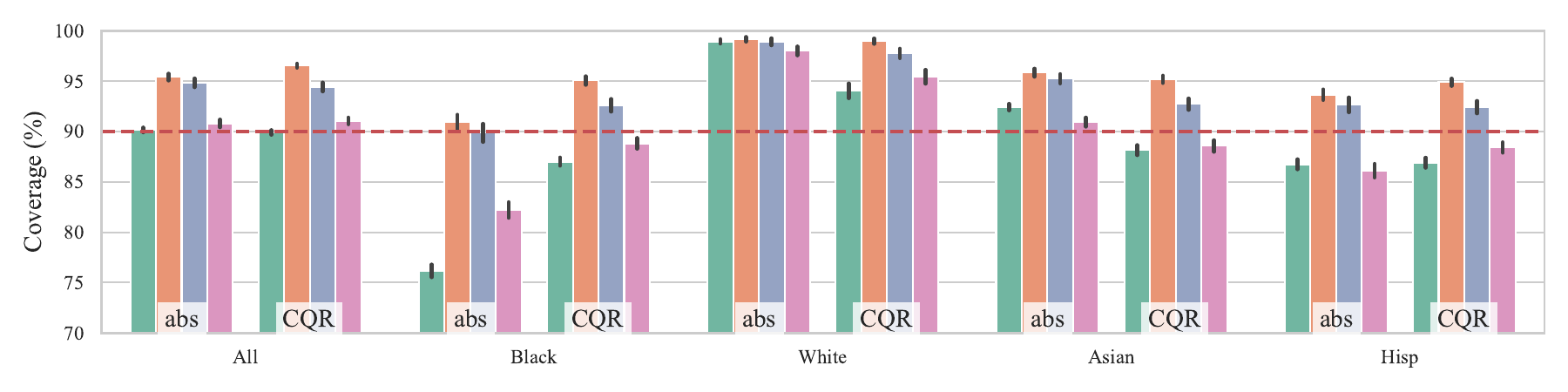} \\ \vspace{-2pt}
    \includegraphics[width=\textwidth]{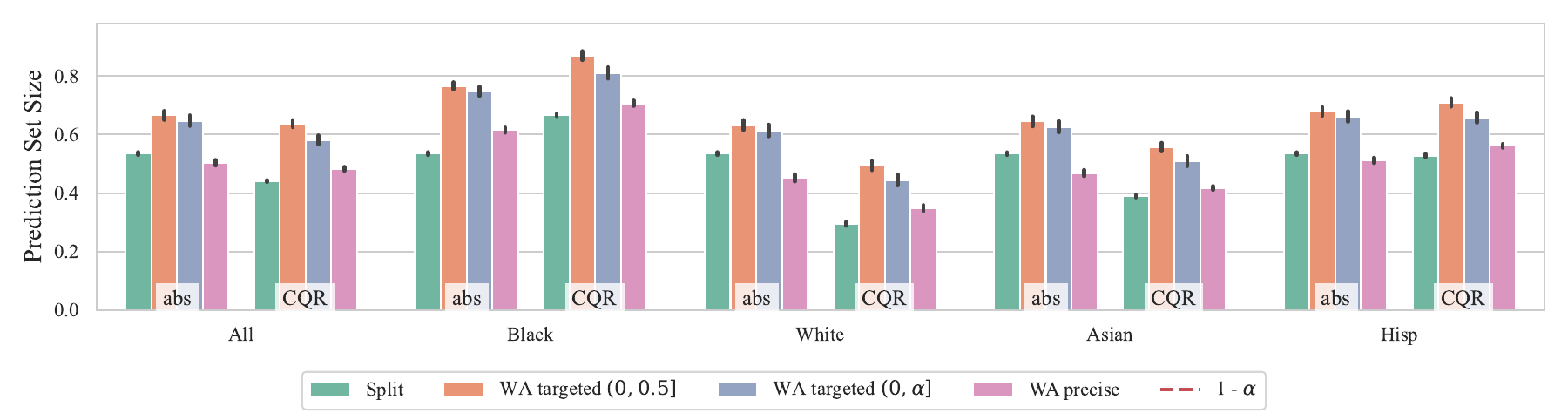} \\ \vspace{-2pt}
    \caption{Coverage (top) and prediction set size (bottom) for split conformal and weighted aggregation variants across subgroups with top 70th percentile racial representation. As in Figure \ref{fig:app_exp_communities_50}, WA targeted meets coverage for all subgroups. Error bars represent 95\% confidence intervals.}
    \label{fig:app_exp_communities_70}
\end{figure}

In this section, we evaluate our methods on the Communities and Crimes dataset \citep{redmond2002data}, where the task is to predict the per capita violent crime rate of a community based on its demographic features, and our primary interest is in understanding how coverage differs across communities with varying racial compositions \citep{gibbs2023conformal}.

Figures \ref{fig:app_exp_communities_50} and \ref{fig:app_exp_communities_70} compare split conformal with weighted aggregation variants in terms of coverage and prediction set size across demographic groups. Unlike Figures \ref{fig:uci_regression_plots} and \ref{fig:uci_classification_plots}, where local validity is assessed via WS coverage, these experiments evaluate local validity in terms of consistency across demographic groups. Thus, we display group-specific performance for each method, which we group further by the two types of nonconformity score, absolute residual and CQR score.

The demographic groups in Figures \ref{fig:app_exp_communities_50} and \ref{fig:app_exp_communities_70} represent communities where a particular racial demographic is in the top $p$-percentile of representation \citep{gibbs2023conformal}. Figure \ref{fig:app_exp_communities_50} shows results for $p = 50$, and Figure \ref{fig:app_exp_communities_70} for $p = 70$.

\paragraph{Marginal coverage and coverage across groups}
The ``All'' category in the coverage plots represents coverage across all demographic groups, or marginal coverage. As in prior experiments, we see that split conformal achieves marginal coverage closest to nominal, with WA precise close behind, while WA targeted variants tend to overcover. However, although split conformal enjoys precise marginal coverage, it also exhibits substantial disparities in performance across demographic groups, significantly undercovering for Black and Hispanic groups while overcovering for White. In contrast, WA variants display far less demographic variation, with WA targeted achieving coverage for all demographic groups.

\paragraph{Comparing nonconformity scores}
For all demographic groups except the Asian group, using CQR scores instead of absolute residuals improves coverage for split conformal, with the improvement being most pronounced for the Black group. However, while CQR reduces undercoverage, it is never sufficient to fully close the coverage gap for an undercovered group. Rather, its primary benefit appears to be in reducing the variability in coverage across groups. Across all methods, the variance in CQR-based coverage is lower than that of absolute residuals, suggesting that CQR contributes to more stable group-wise coverage, even if it does not fully mitigate disparities.

\subsubsection{Comparing to conservative CQR} \label{app:exp_real_conservative_comparison}

In our main results (Figure~\ref{fig:uci_regression_plots}), weighted aggregation typically has higher marginal coverage than CQR (that is, split conformal with CQR nonconformity scores), but also achieves WS coverage that is closer to target. This raises a natural question: does weighted aggregation achieve better WS coverage simply because it is more conservative, or does this reflect genuine adaptivity to difficult regions of the data?

\paragraph{Experimental details} To disentangle these effects, we compare weighted aggregation to a \textit{conservative version of CQR}. For this baseline, we tune the significance level $\alpha$ of CQR so that its empirical marginal coverage matches that of weighted aggregation as closely as possible. (Note that because conformal prediction with a finite calibration set induces a discrete grid of attainable coverage levels, this matching cannot be exact in general; we select the $\alpha$ that yields the closest available coverage.)

\begin{figure*}[t]
    \centering
    \includegraphics[width=\textwidth]{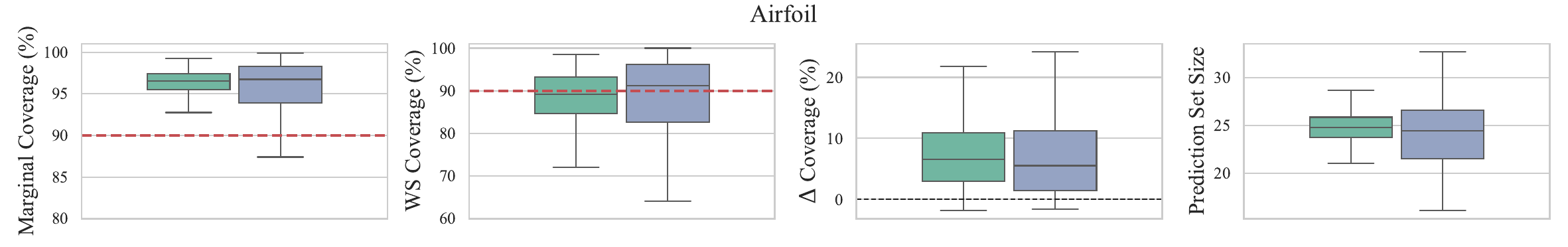} \\ \vspace{-1pt}
    \includegraphics[width=\textwidth]{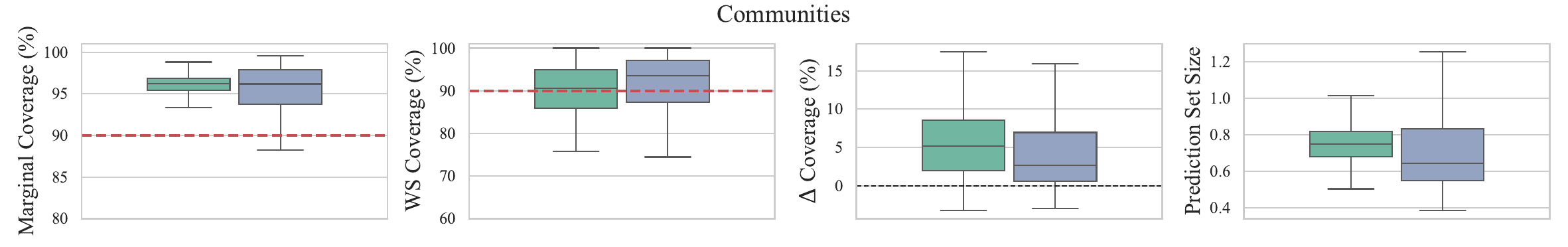} \\ \vspace{-1pt}
    \includegraphics[width=\textwidth]{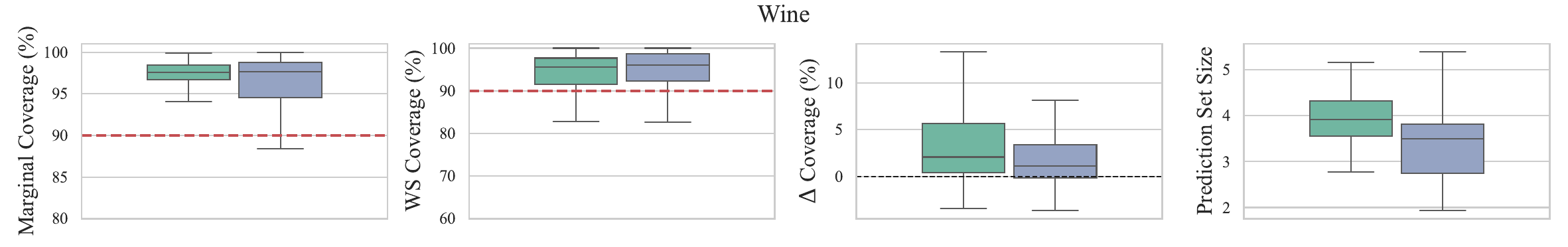} \\ \vspace{-1pt}
    \includegraphics[width=\textwidth]{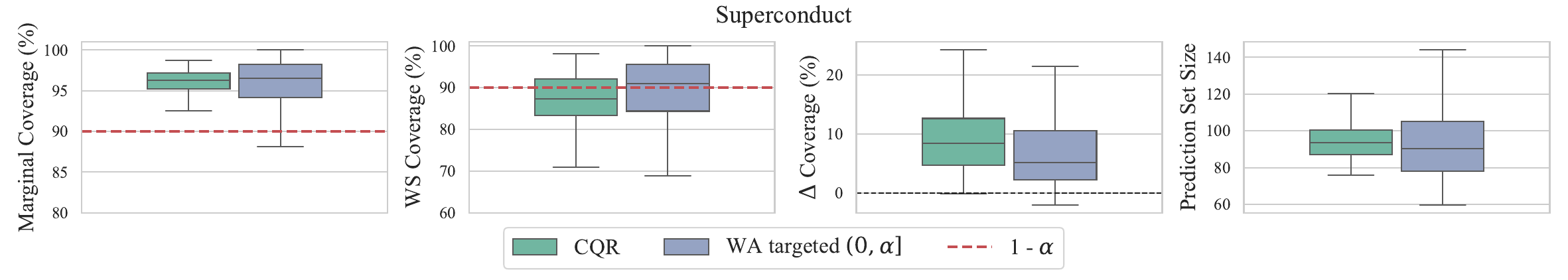} \\ \vspace{-1pt}
    \caption{Comparison between WA targeted $(0, 0.1]$ and CQR at matched empirical marginal coverage. Each row corresponds to a dataset, with plots for marginal coverage, WS coverage, $\Delta$ coverage, and prediction set size from left to right. After matching marginal coverage, WA targeted $(0, 0.1]$ has more uniform coverage (that is, smaller $\Delta$ coverage) and smaller prediction set size on average.
    \vspace{-5pt} 
    }
    \label{fig:uci_conservative_comparisons}
\end{figure*}

We evaluate both methods over 200 trials and summarize the results in Figure~\ref{fig:uci_conservative_comparisons}. To facilitate comparisons even when marginal coverages are not perfectly matched, we also report $\Delta$ coverage, or the difference between the marginal and WS coverage per trial.\footnote{Although potentially unintuitive, empirical WS coverage can occasionally exceed empirical marginal coverage because the WS metric minimizes over a restricted family of slabs (those with at least a $\delta$ fraction of test points, and, in practice, a finite random subset). Consequently, the minimizing slab is not necessarily the region driving undercoverage, and the resulting minimum may be higher than the overall average. For this reason, $\Delta$ coverage can be occasionally less than 0.}

\paragraph{Discussion of Figure \ref{fig:uci_conservative_comparisons}} While matching the marginal coverage does move the WS coverage of conservative CQR closer to that of weighted aggregation, performance differences remain. Conservative CQR still undercovers on the WS slab for the Airfoil and Superconduct datasets, and its average $\Delta$ coverage remains consistently larger across all datasets. This indicates that, even when conservative CQR meets the coverage requirement on average (marginally), it fails to maintain that validity locally (on the WS). The lower $\Delta$ coverage of weighted aggregation shows a smaller gap between average and worst-case coverage, indicating a more uniform distribution of validity across the input space.

Meeting this higher marginal coverage also forces conservative CQR to inflate its prediction sets more than weighted aggregation does. In other words, when forced to be equally conservative overall, CQR improves WS coverage primarily by making sets \textit{globally wider}. In contrast, weighted aggregation improves WS coverage by redistributing uncertainty toward the most challenging regions where coverage failures are more likely. This targeted behavior suggests that weighted aggregation is more adaptive than CQR, rather than simply more conservative.

\clearpage 
\section{Comparison to other aggregation methods} \label{app:other_baselines}

\subsection{Benefits of weighted p-value aggregation} \label{app:other_baselines_benefits}
Weighted p-value aggregation is particularly effective in the realistic and practically important scenario when the predictors to be aggregated are not perfectly aligned: that is, when the different predictors capture complementary aspects of a problem, so that no single predictor is uniformly superior (more predictive) across the input space. This setting is represented in Figure \ref{fig:summary_figure}, and can occur, for example, when the predictors are trained on different modalities or specialize in different operating conditions. In this scenario, an effective aggregation method should (1) faithfully reflect whether the predictors agree or disagree, and (2) adjust to a test point by assigning more weight to the most relevant predictors at that point.

In the standard MoE formulation, weights are optimized for predictive accuracy (under mean squared error or cross-entropy loss), and reflect how relevant each expert is for a given test point. Under this setting, weighted p-value aggregation achieves both of these goals. We elaborate on this below.

\subsubsection{p-value aggregation best reflects expert disagreement}
One way to taxonomize aggregation methods is to consider whether they combine \textit{calibrated} uncertainty representations (like p-values or prediction sets) or \textit{pre-calibrated} representations (like scores). To illustrate how these differ, consider the extreme case of expert diversity where two experts have disjoint prediction sets.

\paragraph{Pre-calibration aggregation methods eliminate disagreement}
Methods that combine scores before calibration \citep{luoconformity} convert the experts’ diverse behaviors into a single blended score that no longer encodes which expert supported which region. In the disjoint expert example, averaging pre-calibration scores would collapse the final prediction set into a single region that neither expert truly supports.

To see this more explicitly, let each expert use a standard convex conformal score (e.g. absolute residual or quantile regression score) so each $s_k(x,y)$ is convex in $y$. Then, the average score $s_{\text{avg}}(x,y) = \sum_k w_k s_k(x,y)$ is convex, and the conformal set produced from this score is the sublevel set $C_{\text{avg}} = \{ y: s_{\text{avg}}(x,y) \le Q_\alpha^+ \}$, and is also convex. Thus, even if the experts’ individual prediction sets are disjoint, the averaged score produces one connected prediction set which may not reflect the support of either expert.

\paragraph{Set selection methods eliminate disagreement}
Methods that select one prediction set \citep{yang2024selection} necessarily discard other experts’ predictions. In the disjoint expert example, the aggregate set reflects only one expert’s prediction, and all complementary evidence from the other expert is lost. This means the aggregate set ignores valid, calibrated support for labels that remain plausible.

\paragraph{p-value aggregation preserves the degree of support from each expert}
In the end, the only methods that preserve expert disagreement are the methods that combine, and do not just select, \textit{calibrated} representations---p-value aggregation, majority vote \citep{cherubin2019majority}, and set union. However, these methods differ substantially in how much expert information they retain.

Majority vote and set union operate on binary membership. For weighted majority vote \citep{gasparin2024merging}, the prediction set is
\begin{equation*}
    C_{\text{WMV}}(x) = \left\{ y:\sum_k v_k \mathds{1} \{ p_k(x,y) > \alpha \} \ge \frac{1}{2} \right\},
\end{equation*}
so each expert contributes only the indicator $\mathds{1} \{ p_k(x,y) > \alpha \}$. This tracks whether an expert supports a label, but not how strongly they support it. In contrast, p-value aggregation combines the full p-value functions, with prediction set
\begin{equation*}
    C_{\text{p-value}}(x) = \left\{ y : \sum_k w_k p_k(x,y) > \alpha \right\}
\end{equation*}
(restated from \eqref{eq:ps_avg}). If the $i$th expert assigns $p_i(x,y)=0.95$ and the $j$th expert assigns $p_j(x,y)=0.12$ at $\alpha=0.1$, majority vote records only that both experts include $(x,y)$, while p-value aggregation keeps their relative strength with $w_i\cdot 0.95 + w_j\cdot 0.12$. Thus, p-value aggregation \textit{preserves graded expert information}, and can distinguish regions of strong support from regions of weak support. This sensitivity to the degree of support from each expert allows p-value aggregation to preserve disagreement more faithfully than binary set-based methods.

\paragraph{Why preserving disagreement matters}
With complementary experts, disagreement often reflects genuine multi-modality or heterogeneity in the underlying data. Aggregation methods that collapse this structure risk including labels no expert supports, or excluding labels at least one expert favors. By retaining the full shape and strength of each expert’s p-value function, p-value aggregation ensures that the final set aligns more closely with the true uncertainty represented across experts.

\subsubsection{p-value aggregation allows for data-dependent weights}
As we see in \eqref{eq:p-variable}, p-variables satisfy a direct and explicit validity condition. This condition is expressed entirely in terms of the distribution function of a single random variable, and as such, is \textit{manipulable}: we can combine p-values, examine how the combination behaves relative to this inequality, and apply a correction that restores validity. Prediction sets and nonconformity scores do not admit a similarly explicit validity criterion that allows the aggregated object to be repaired directly after reweighting, so no analogous simple correction exists for majority vote, set union, or score-based averaging.

Consider again the case where experts specialize in different regions of the input space, so that the expert that is most relevant at each point varies across the domain. Majority vote and set union cannot express this form of adaptivity. Even in weighted majority vote, the weights must be fixed in advance or otherwise independent of the test input, so the weighting rule is applied uniformly across the input space. This means every expert contributes its vote at every test point in the same way, regardless of whether that expert is informative for the local region. When an expert is poorly suited to the region around the test point, its vote can (at best) widen the prediction set unnecessarily, or even (at worst) counteract the experts that are actually reliable there.

Set selection methods have a more fundamental limitation. By construction, set selection yields a single global choice of expert, and this remains fixed for all future test points. This is incompatible with any notion of input adaptivity, which is important to settings where local conditions change.

\subsection{Comparison to additional adaptive baselines} \label{app:other_baselines_adaptive}
\paragraph{Additional adaptive baselines}
One of the benefits of our proposed method is that its extension to data-dependent weights allows for the weight-aggregated set to adapt to the specific test input. This property of adapting to the test input is called \textit{local}, \textit{adaptive}, or \textit{conditional}.

In our experiments in \S\ref{sec:moe}, we compare to conformal quantile regression (CQR) \citep{romano2019conformalized} as our adaptive baseline. CQR is simple but powerful, and can be considered a seminal adaptive method with multiple follow-up works \citep{sesia2020comparison, liu2022conformalized, jensen2022ensemble, rossellini2024integrating}. Our experiments in \S\ref{sec:moe} both compare to CQR (represented as split conformal with CQR scores), and combine our method with CQR (represented as weighted aggregation with CQR scores). To further contextualize the local validity of our approach, we now compare to two additional baselines.    
\begin{itemize}
    \item \textit{Localized conformal prediction (LCP)} \citep{guan2019conformal}: LCP reweights calibration scores with weights $w_i \propto H(X_i, X_{n+1})$, where the localization kernel $H$ measures the proximity between the test point $X_{n+1}$ and each calibration point $X_i$. At test time, LCP forms an empirical distribution of weighted calibration scores and solves for the significance level $\tilde{\alpha}(x_{n+1})$ that restores marginal coverage after the pointwise weighting. LCP can be computationally intensive, as each test point requires kernel weights over all calibration points.
    \item \textit{Locally weighted conformal prediction (LWCP)} \citep{lei2018distribution}: LWCP normalizes calibration residuals by fitting one model for the mean, a second model for the local error scale, and then rescaling each calibration residual by its estimated variability. Running conformal prediction on these normalized scores yields intervals whose widths adapt to local uncertainty.
\end{itemize}
We focus on comparing to methods that improve coverage across the input space in a general data-adaptive way. For this reason, we do not compare to methods whose local validity requires the user to first specify a target family of subpopulations or distributional shifts, like \citet{gibbs2023conformal}.

\paragraph{Weighted aggregation can be combined with other adaptive methods} LCP introduces adaptivity by reweighting calibration scores, and CQR and LWCP introduce adaptivity through adaptive score functions. In contrast, our method introduces adaptivity at the \textit{aggregation} step, by reweighting the post-calibration p-value functions in a data-dependent way.

When these adaptive methods operate on different steps of the conformal procedure, they generally do not interfere with one another. Thus, because weighted aggregation introduces adaptivity at a different step than LCP, CQR, or LWCP, it complements rather than competes with these methods, and can in principle be combined with any of them.

\begin{figure*}[t]
    \centering
    \includegraphics[width=\textwidth]{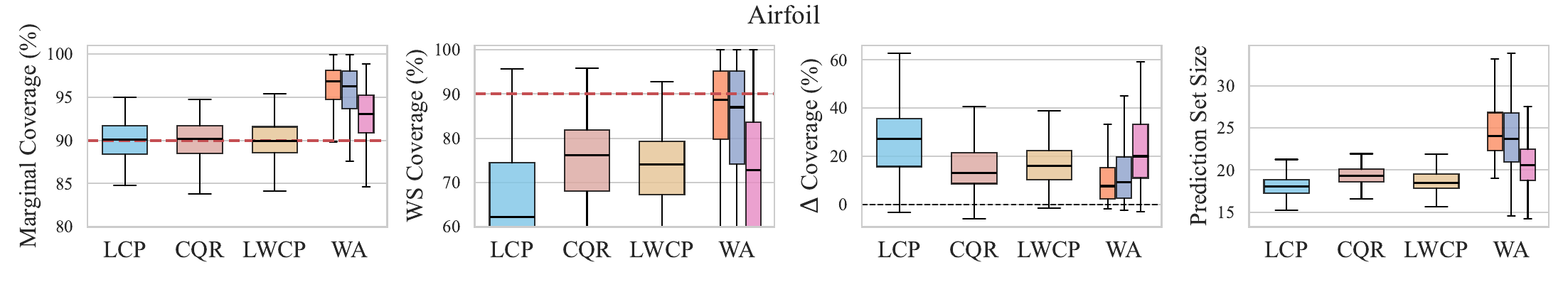} \\ \vspace{-1pt}
    \includegraphics[width=\textwidth]{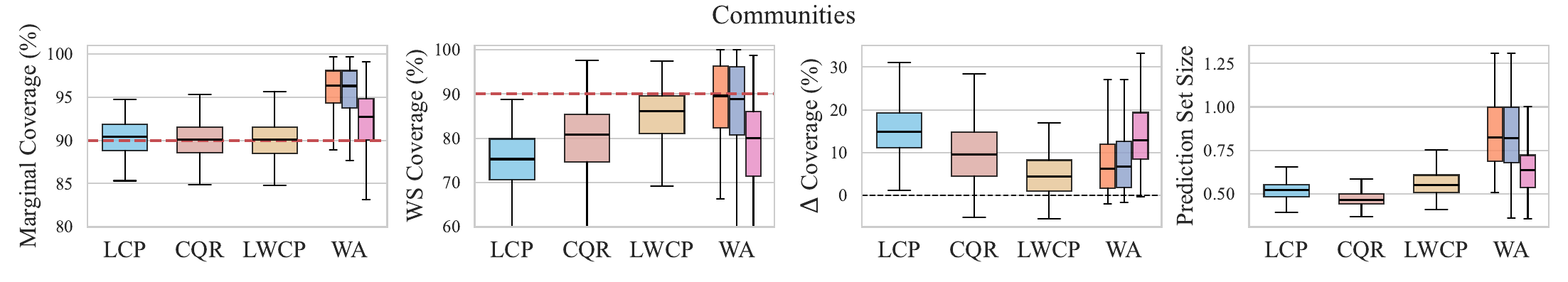} \\ \vspace{-1pt}
    \includegraphics[width=\textwidth]{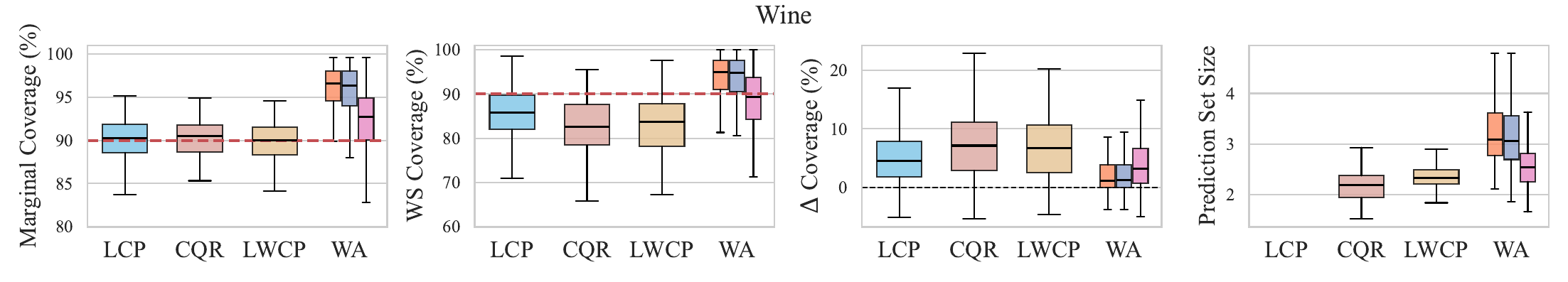} \\ \vspace{-1pt}
    \includegraphics[width=\textwidth]{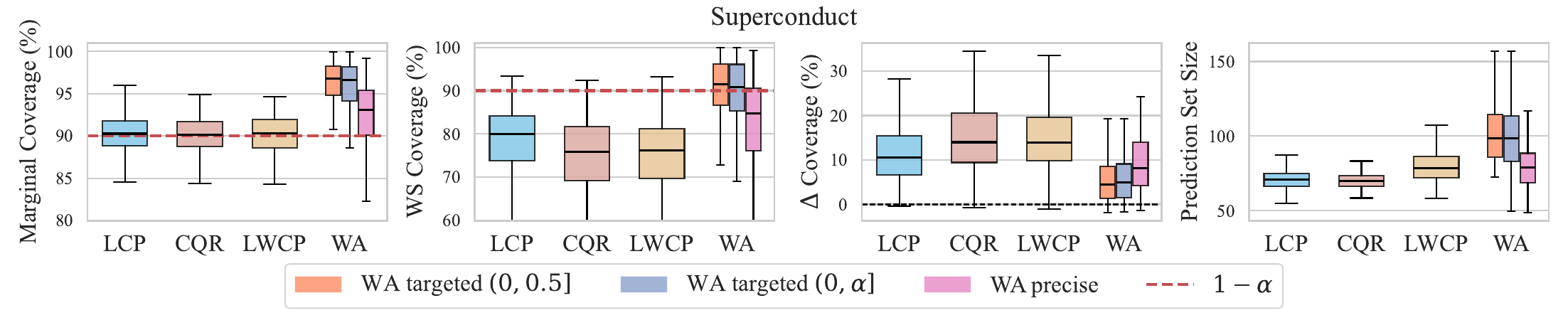} \\ \vspace{-1pt}
    \caption{Comparison between weighted aggregation and adaptive baselines. Each row corresponds to a dataset, with plots for marginal coverage, WS coverage, $\Delta$ coverage, and prediction set size from left to right. Although the weighted aggregation methods (WA) are conservative marginally, it also has the most uniform coverage (that is, the smallest $\Delta$ coverage overall), and achieves closest-to-nominal coverage on the WS slab.
    \vspace{-5pt} 
    }
    \label{fig:uci_adaptive_comparisons}
\end{figure*}

\paragraph{Experimental details} For this experiment, we run each of the methods separately to isolate their individual effects. These results are therefore presented differently than in Figure \ref{fig:uci_regression_plots}, where we had to visualize composition between the absolute residual score / CQR score and split conformal / weighted aggregation. In Figure \ref{fig:uci_adaptive_comparisons}, LCP, CQR, and LWCP are separate methods that are each compared to the weighted aggregation methods (WA) directly. (More precisely, LCP, CQR, and LWCP are each implemented on top of split conformal with the full MoE predictor---each method replacing only the component it introduces, with the rest of the procedure matching split conformal using the full MoE.)

We evaluate all methods over 200 trials and report distributions of marginal coverage, WS coverage, $\Delta$ coverage, and prediction set size (as in Figure~\ref{fig:uci_conservative_comparisons}). Figure~\ref{fig:uci_adaptive_comparisons} summarizes the results. 

\paragraph{Discussion of Figure \ref{fig:uci_adaptive_comparisons}} 
The comparisons between weighted aggregation and the adaptive baselines reflect a trade-off between marginal efficiency and adaptivity. Weighted aggregation is marginally overconservative, whereas the other methods have marginal coverage close to target. However, the weighted aggregation methods consistently brings WS coverage close to target, while the other methods substantially undercover the WS slab. In addition, weighted aggregation almost always achieves the best $\Delta$ coverage: even though its marginal coverage is conservative, the gap between marginal and WS is almost always smaller than the other baselines, indicating that weighted aggregation reallocates uncertainty more effectively in the hardest regions (the WS slab). These results suggest that the main advantages of weighted aggregation are in its \textit{adaptivity}, by providing more uniform coverage overall and achieving better coverage on the WS slab. We therefore suggest to use weighted aggregation when performance on challenging/minority slices is a priority.

\subsection{Weighted majority vote} \label{app:wmv}
The idea to combine conformal prediction sets by weighted majority vote was introduced by \citet{gasparin2024merging} as an extension of the majority vote method first proposed by \citet{cherubin2019majority}. At first consideration, weighted majority vote appears to differ from our method: weighted majority vote performs weighted aggregation of \textit{prediction sets}, while our method performs weighted aggregation of the \textit{p-values associated with prediction sets}. Despite this distinction, Appendix B of \citet{gasparin2024merging} observes that these two methods are, in fact, dual to each other under data-independent weights, the setting considered in their work. Nevertheless, our p-value formulation enables two key extensions that go beyond weighted majority vote.
\begin{itemize}
    \item Our formulation allows us to apply the result of \citet{vovk2020combining} to strengthen coverage guarantees for \textit{data-independent} weights when they are sufficiently asymmetric.
    \item Our formulation provides a principled extension to \textit{data-dependent} weights by transforming the weighted average of p-variables to also be a valid p-variable. Not only does this allow us to use weights learned from data, but it also yields a form of local validity, a property not available to existing set aggregation methods.
\end{itemize}
Because our method is a dual formulation to the weighted majority vote method of \citet{gasparin2024merging}, we do not include it as a separate baseline to avoid redundancy.

\subsection{\texorpdfstring{Extending the p-variable transformation of \citet{stutz2023conformal} to weighted aggregation}{Extending the p-variable transformation of Stutz et al. [30] to weighted aggregation}} \label{app:comparison_transformation_stutz}
To the best of our knowledge, weighted majority vote \citep{gasparin2024merging, gasparin2024conformal} is the only existing method to address weighted prediction set aggregation, and the method of \citet{stutz2023conformal} is designed for the different problem of uncertainty in the ground truth labels. To address their problem, \citet{stutz2023conformal} propose sampling $m$ labels for each calibration point and using the labels to compute $m$ p-values, then taking the unweighted average of these p-values and applying a transformation to obtain a valid p-variable.

Although \citet{stutz2023conformal} only consider the unweighted average, their transformation is general enough to apply to a weighted average of p-values as well, and can therefore be adapted to our setting. 

\paragraph{How the transformations affect the weights}
Both our method and the method of \citet{stutz2023conformal} aim to transform a random variable to a p-variable to maintain coverage guarantees. The difference between the two methods lies in the nature of the transformation. Our method applies a \textit{linear} transformation that preserves the proportions of the weights; this can be important when the weights reflect meaningful quantities, like the weights learned by the routing network of an MoE model. In contrast, \citet{stutz2023conformal} apply a \textit{nonlinear, rank-based} transformation by computing the empirical CDF of the random variable and returning its value at the observed point. That is, given a random variable $X$, their method estimates its CDF $F$ and uses $F(X)$ as the resulting p-value. While this guarantees validity and preserves ordering, it does not preserve the relative scale between values and therefore discards some of the semantics of the original weights.

\paragraph{Comparing both transformations with our MoE setup} 
In our method for weighted prediction set aggregation with data-dependent weights, we transform the weighted average of the individual prediction set p-values to a valid p-variable using a linear scaling. This transformation allows us to maintain a coverage guarantee for the prediction set defined by the weighted average of the p-values. In the MoE setting, this prediction set corresponds to aggregating the p-values from each expert according to the weights learned by the routing network.

To adapt the method of \citet{stutz2023conformal} to this context, we substitute their transformation in place of our linear scaling. We now restate their original setting and transformation in more detail to clarify how their method can be extended to our setting. 

The transformation proposed by \citet{stutz2023conformal} was originally developed to address the problem of uncertainty in the ground truth labels. In their setting, each calibration point consists only of an input \(X_i\); they sample \(m\) labels to get \(m\) p-values for each calibration point, and then take the unweighted average of these p-values. We denote this unweighted average as \(P_{\text{avg}}^i\), with distribution function \(F\). Their method transforms \(P_{\text{avg}}^i\) into a valid p-value via \(F(P_{\text{avg}}^i)\), and the prediction set is then the set of all labels such that the corrected p-value \(F(P_{\text{avg}}^i)\) corresponding to each sample is greater than some threshold \(\alpha\). 

The coverage guarantee for this corrected p-value holds only with an asymptotic number of samples, as the empirical CDF \(\widehat{F}\) approaches the true CDF \(F\). To establish a finite-sample guarantee, \citet{stutz2023conformal} introduce a DKW-derived correction \(\epsilon\) and define their prediction sets based on \(\widehat{F} + \epsilon\). We find that, although their proposed prediction set yields an elegant \((1-\alpha)(1-\delta)\) finite-sample guarantee, the \(\epsilon\) correction is extremely conservative in practice.

Let us refer to the finite-sample empirical CDF method as \textit{ECDF-DKW}, and the variant without the DKW correction as simply \textit{ECDF}. We now present additional experiments where we recreate the main findings of our paper with the ECDF and ECDF-DKW methods.

\begin{figure}[t]
    \centering
    \includegraphics[width=\textwidth]{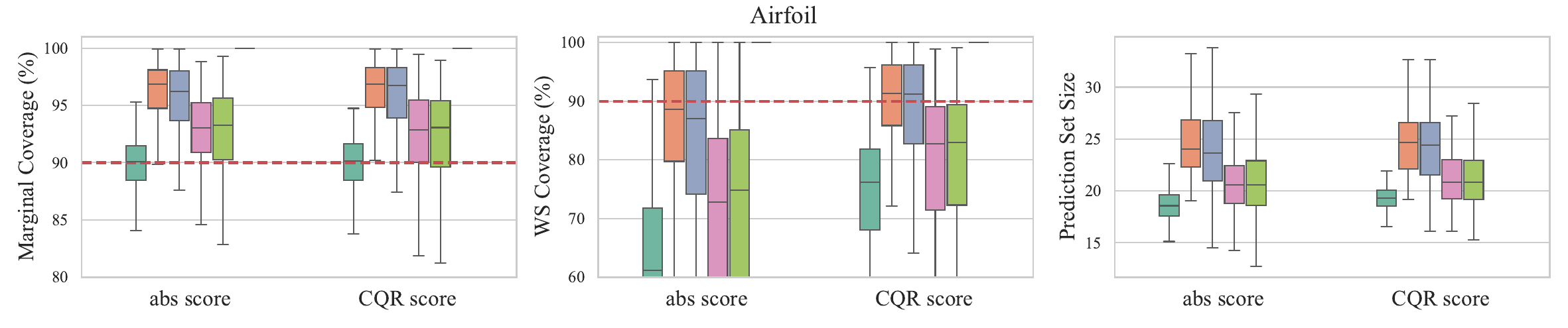} \\ \vspace{-2pt}
    \includegraphics[width=\textwidth]{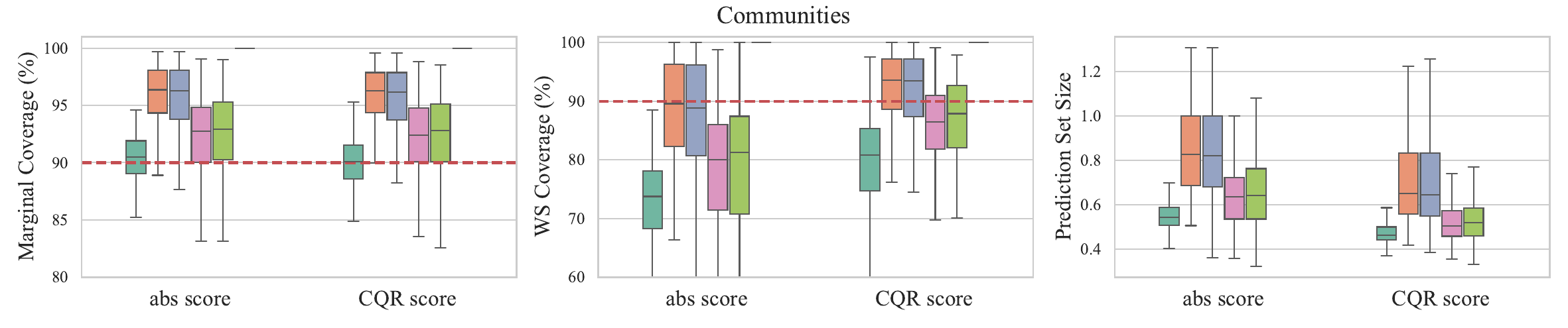} \\ \vspace{-2pt}
    \includegraphics[width=\textwidth]{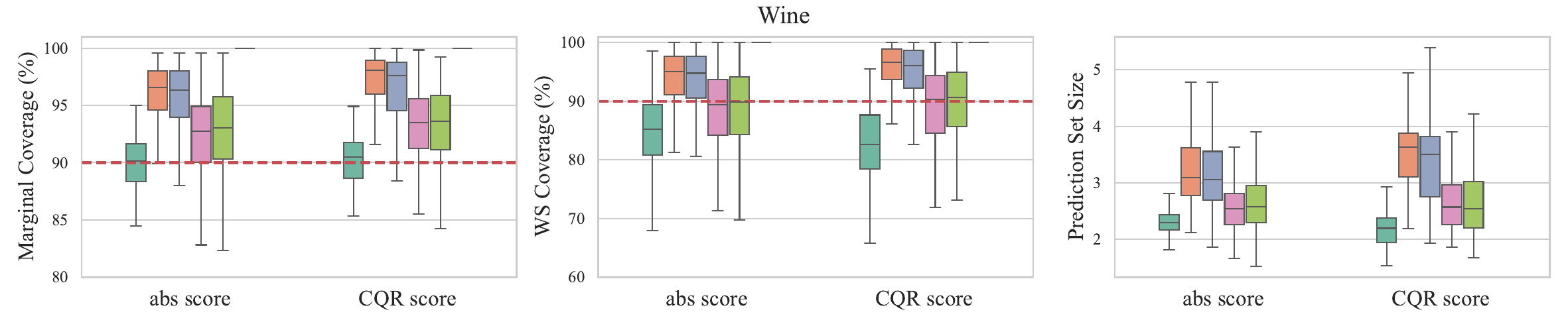} \\ \vspace{-2pt}
    \includegraphics[width=\textwidth]{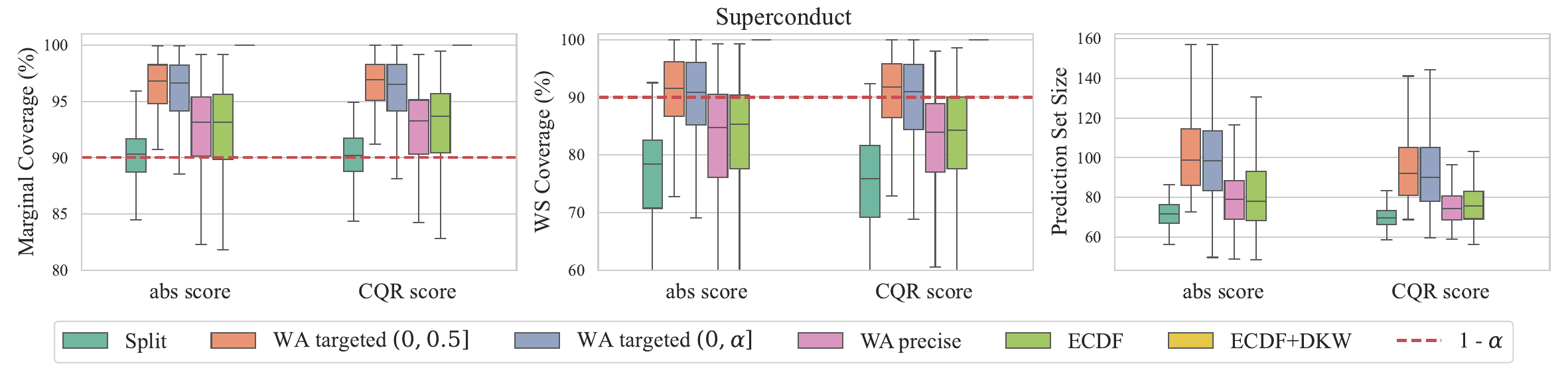} \\ \vspace{-2pt}
    \caption{Regression experiments of our linear transformation method with ECDF and ECDF-DKW. Each row corresponds to a dataset, with plots for marginal coverage, WS coverage, and prediction set size from left to right. ECDF performs similarly (sometimes slightly more conservatively) to WA precise in terms of coverage and efficiency. However, ECDF-DKW is so conservative that it covers the entire label space.
    \vspace{-10pt} 
    }
    \label{fig:stutz_regression}
\end{figure}

Figure \ref{fig:stutz_regression} recreates the regression experiments of Figure \ref{fig:uci_regression_plots}, with the addition of ECDF and ECDF-DKW. We note that ECDF performs very similarly to WA precise in terms of coverage (left), WS coverage (middle), and prediction set size (right), with ECDF being slightly more conservative in most cases. Like WA precise, ECDF is more conservative than split conformal and less conservative than WA targeted, although it still often falls short of the \(1-\alpha\) guarantee on the WS slab. On the other hand, ECDF-DKW is so conservative that its prediction sets cover the entire label space. We represent this with 100\% coverage on both the coverage plots and WS coverage plots, and we omit ECDF-DKW from the prediction set size plots.

\begin{figure}[t]
    \centering
    \includegraphics[width=\textwidth]{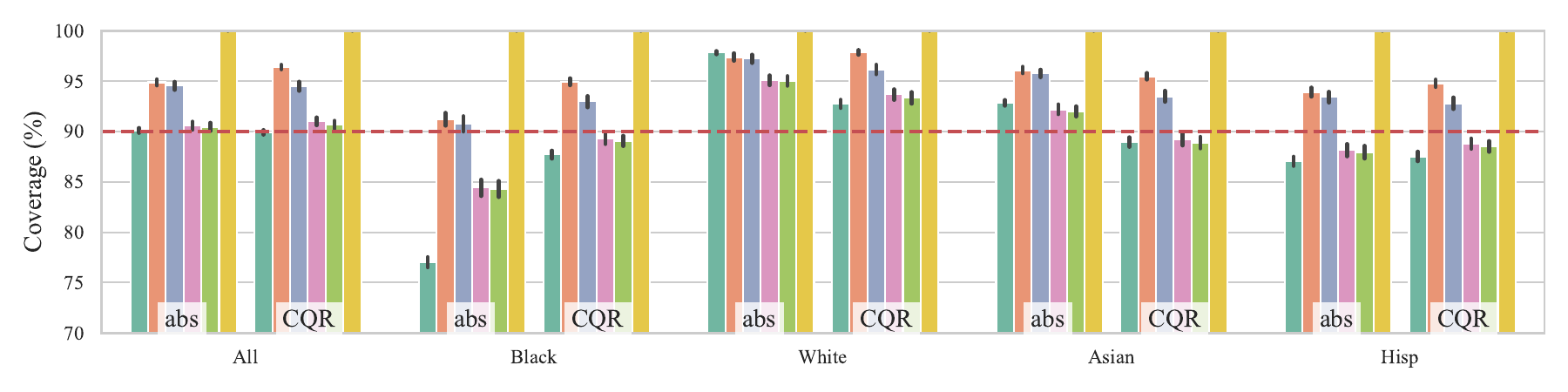} \\ \vspace{-2pt}
    \includegraphics[width=\textwidth]{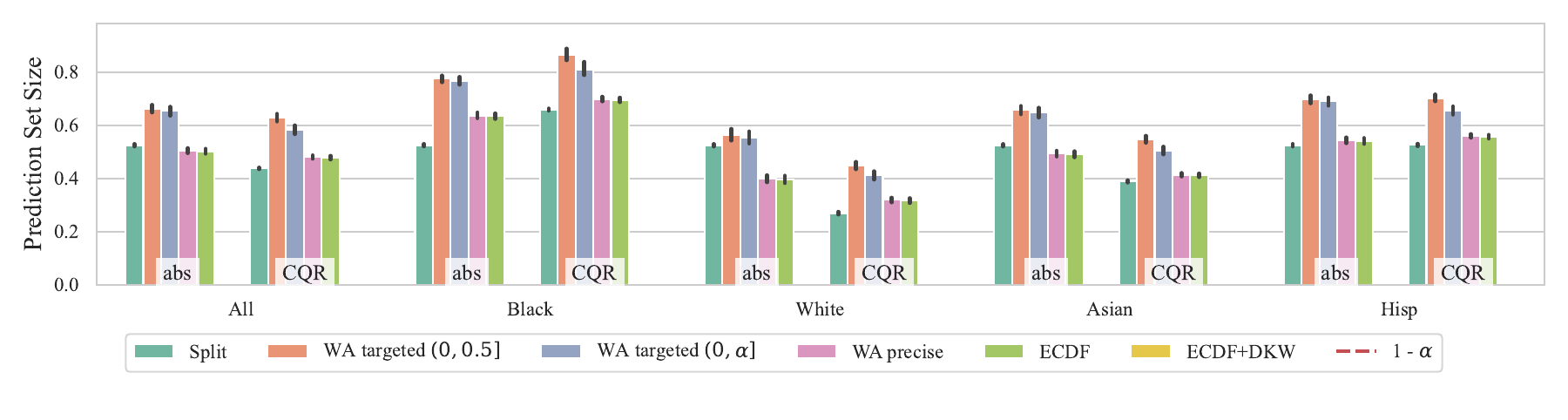} \\ \vspace{-8pt}
    \caption{Coverage (top) and prediction set size (bottom) for subgroups with top 50th percentile racial representation in the UCI Communities and Crimes dataset. ECDF performs similarly to WA precise; ECDF-DKW is so conservative that it covers the entire label space. Error bars represent 95\% confidence intervals.
    }
    \label{fig:stutz_communities}
\end{figure}

Figure \ref{fig:stutz_communities} recreates the Communities and Crimes experiment of Figure \ref{fig:app_exp_communities_50} with ECDF and ECDF-DKW. Again, we see that ECDF performs very similarly to WA precise on our demographic-conditioned view of Communities and Crimes, with similar coverage (top) and prediction set size (bottom) across all demographics---with ECDF having slightly lower coverage on most demographics, including demographics where WA precise undercovers. As before, we observe that ECDF-DKW is so conservative that it has 100\% coverage and unbounded prediction sets.

\begin{figure}[t]
    \centering
    \includegraphics[width=0.6\textwidth]{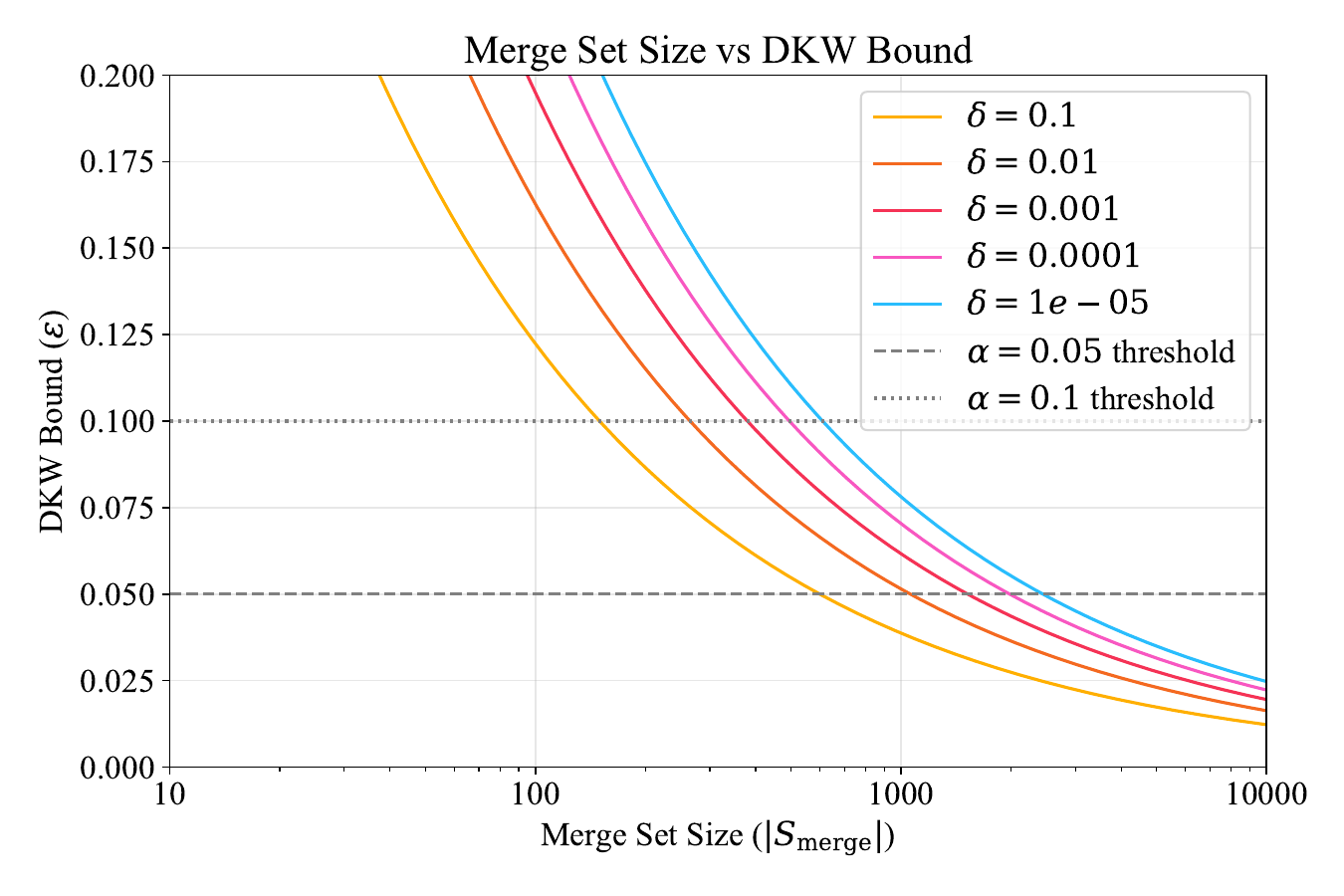}
    \vspace{-5pt}
    \caption{Finite-sample correction \(\epsilon\) used in ECDF-DKW as a function of the merging set size \( |S_{\text{merge}}| \) and user-specified significance level \(\delta\). This correction is the offset required to ensure the finite sample guarantee of \((1-\alpha)(1-\delta)\) in \citet{stutz2023conformal}. Dashed lines mark $\alpha$ levels of 0.05 and 0.1. When \(\epsilon > \alpha\), the prediction set includes all labels and becomes unbounded, yielding 100\% coverage.}
    \label{fig:dkw}
\end{figure}

\paragraph{Why is ECDF-DKW so conservative?}
For ECDF-DKW, the finite-sample variant of ECDF, \citet{stutz2023conformal} use DKW to add a finite-sample correction \(\epsilon\) to the empirical CDF \(\widehat{F}(P_{\text{all}})\), then compare this sum to the significance level \(\alpha\). The prediction set with finite-sample guarantees is therefore the set of all labels such that \(\widehat{F}(P_{\text{avg}}) + \epsilon > \alpha\). However, if \(\epsilon\) is already greater than \(\alpha\), then this condition is always satisfied and the prediction set includes \textit{all} labels, becoming unbounded.

The finite-sample correction \(\epsilon\) is a function of the number of samples used to compute the empirical CDF. Figure \ref{fig:dkw} shows that when the merging set size is less than 1000, \(\epsilon\) is typically large enough to exceed common values of \(\alpha\), making unbounded sets very likely.

\newpage
\section{A note on computational complexity}  \label{app:complexity}
The computational complexity of our method matches the complexity of existing prediction set aggregation methods for the case of data-independent weights, and includes an additional one-time cost to compute an empirical CDF for the case of data-dependent weights.

The complexity of prediction set aggregation methods was first observed by \citeauthor{cherubin2019majority}, who compares the computational overhead of their majority vote aggregation method with the overhead of p-value aggregation methods. They note that while majority vote requires simpler operations to determine whether each label is included in the final prediction set, it also requires that predictions be recomputed for each significance level, making it less efficient when sets must be constructed at multiple thresholds. In contrast, p-value methods allow the aggregation to be computed once and then applied to any significance level without additional computation.

These trade-offs in speed and cost may influence which method is better suited to a given application---for example, p-value aggregation may be preferable if prediction sets need to be dynamically thresholded at test time. However, the time complexity of both methods is the same: combining \( K \) prediction sets for \( N \) test objects with a label space size of \( L = |\mathcal{Y}| \) has complexity \( \mathcal{O}(KLN) \). This complexity is necessary for all prediction set aggregation methods, as it reflects the cost of evaluating multiple prediction sets across the label space.

With data-independent weights, our method matches this \( \mathcal{O}(KLN) \) complexity directly. With data-dependent weights, the only additional computation required is a one-time estimation of a correction factor \( \widehat{m}^* \) from data split \( S_{\text{merge}} \). This step involves computing the empirical CDF of the weighted average p-values on \( S_{\text{merge}} \), which has a complexity of \( \mathcal{O}(M \log M) \) for a split of size \( M \). Importantly, this correction is computed once and does not require retraining, and the rest of the procedure for data-dependent weights has the same \( \mathcal{O}(KLN) \) cost as other prediction set aggregation methods.

\newpage
\section{Further implementation details} \label{app:exp_details}

\subsection{Data} \label{app:exp_details_data_splits}
\paragraph{Synthetic data} \label{app:exp_details_synthetic}
For the ablation/expository experiments in \S\ref{app:exp_synthetic}, we generate a simple homoskedastic dataset to simulate a regression task. Each input is a 16-dimensional vector drawn from a standard normal distribution, and the output label is the sum of the feature values with additive Gaussian noise. Specifically, for each sample $X_i$, we have
\begin{equation*}
y_i = \sum_{j=1}^{16} X_{ij} + \varepsilon_i,
\end{equation*}
where $\varepsilon_i \sim \mathcal{N}(0, \sigma^2)$ represents additive noise with standard deviation $\sigma = 0.1$. We generate both training and test datasets by independently drawing samples and computing the corresponding target labels.

\paragraph{Data split}
In all experiments, we randomly partition the data into four disjoint subsets of train, calibration, merge, and test. For the experiments on UCI data, we allocate 400 samples for train/calibration/merge with a 50/40/10 split (where the number of training samples follows \citet{barber2021predictive}). For methods that do not require a merge set (e.g. split conformal), we use a 50/50 train/cal split over the same 400 sample budget. This ensures that all methods use the same amount of training data.

\subsection{Mixture-of-experts (MoE) model and optimization} \label{app:exp_details_models_training}
\paragraph{Model architecture}
To isolate the effects of aggregation from model complexity, we use linear models for both the experts and the routing networks.

We define an MoE model with $K$ experts, where the $i$th expert is a linear model $f_i(x; \theta_i)$ parameterized by $\theta_i$, and the routing network is a linear model parameterized by $\phi$. For each expert $i$, it produces a raw output $g_i(x; \phi)$. To ensure the expert weights $\{w_i(x)\}_{i=1}^{K}$ are positive and sum to one, we apply a softmax function to these outputs:
\begin{equation*}
    w_i(x) = \frac{\exp(g_i(x; \phi))}{\sum_{j=1}^{K} \exp(g_j(x; \phi))},
\end{equation*}
These weights determine the contribution of each expert to the final prediction.

\paragraph{Feature assignments} 
In the UCI experiments, the features for each dataset are partitioned into semantically related groups, and each expert specializes in a single group (see \S\ref{app:exp_details_uci_features} for feature groups). In our ablation experiments on synthetic data, we vary the degree of feature overlap between experts to explicitly study its impact on coverage in \S\ref{app:synthetic_feature_assignment}.

\paragraph{Optimization and sequential training} All models are optimized using L-BFGS to minimize mean squared error (MSE). We adopt a sequential training procedure to improve stability \citep{nie2021evomoe, lin2024moe}. First, we train each expert independently on its assigned subset of features. Then, we train the routing network with the full MoE to minimize MSE of the aggregated prediction
\begin{equation*}
    \ell_{\text{routing}} = \frac{1}{|S_{\text{train}}|} \sum_{i \in S_{\text{train}}} \left( y_i - \sum_{j=1}^{K} w_{j}(x_i) f_j(x_i; \theta_j) \right)^2.
\end{equation*}
Note that, while experts may observe different features, all components are trained on the same train set.

\paragraph{Routing at inference}
In our experiments, the trained MoE model is utilized in two ways. For split conformal, we use the weighted sum of experts as a standard black-box predictor. For weighted aggregation, we use the weights from the routing network to scale the p-value functions of each expert.

\subsection{Implementation of weighted aggregation} \label{app:exp_details_WA}
\paragraph{Computing prediction sets}
We leverage the structure of the p-value function to efficiently compute prediction sets.

For a fixed input $x$, the function $\widehat{p}_k(x,y)$ \eqref{eq:pk} is piecewise constant, with discontinuities occurring only at values of $y$ that solve $s_k(x,y) = s_k(X_i,Y_i)$. (For an absolute residual score function $s_k(x,y) = |y-\widehat{\mu}_k(x)|$, these values correspond to $\widehat{\mu}_k(x) \pm s(X_i, Y_i)$ for all $i \in S_k$.) Thus, for each test point $x$, $\widehat{p}_k$ is a step function of $y$ with finite discontinuities, and the weighted average p-value function is also a step function of $y$, with its discontinuities as a union of the $K$ separate sets of discontinuities. We compute prediction sets by identifying these candidate discontinuities, evaluating the p-value at the midpoint of each interval, then taking the union of all intervals where $\widehat{p}(x,y) > \alpha$.

We use the \texttt{portion} library \citep{portion} to represent prediction sets and efficiently compute their Lebesgue measures without discretization.

\paragraph{Computing correction factors $\widehat{m}^*$, $\widehat{m}^\dagger$, $\widehat{m}^\ddagger$}
All correction factors are computed using the empirical CDF of the aggregated p-values on the merging set $S_{\text{merge}}$.

For each point $(X_i, Y_i) \in S_{\text{merge}}$, we calculate the aggregate p-value function $P_{\text{all}}^i = p_{\text{all}}(X_i, Y_i; \mathbf{W}^{(i)})$. We then construct the empirical CDF $\widehat{F}$ and compute the ratios $\widehat{F}(P_{\text{all}}^i) / P_{\text{all}}^i$ for all points in the set.
\begin{itemize}
    \item $\widehat{m}^*$ is the maximum of these ratios over $i \in S_{\text{merge}}$.
    \item $\widehat{m}^\dagger$ (targeted) restricts this maximum to points where $\widehat{F}(P_{\text{all}}^i) \le \alpha'$.
    \item $\widehat{m}^\ddagger$ (precise) uses only the first point $i$ where $\widehat{F}(P_{\text{all}}^i) \ge \alpha'$.
\end{itemize}
Formal definitions for $\widehat{m}^*$ and $\widehat{m}^\dagger$ are provided in \S\ref{app:computing_mhatstar}.

\paragraph{Role of the merging set $S_{\text{merge}}$}
The merging set is used to construct an empirical CDF in order to estimate the correction factor $m^*$ that makes the aggregated p-value valid. The size of this set induces a trade-off between giving a \textit{conservative} or \textit{precise} estimate, which we demonstrate experimentally in \S\ref{app:synthetic_exp_smerge}.

A small $|S_{\text{merge}}|$ results in a noisier empirical CDF; to maintain validity, the correction factor tends to be large, leading to overconservative prediction sets. As $|S_{\text{merge}}|$ increases, the empirical CDF becomes closer to the true CDF, allowing for a more precise estimate of $m^*$ and coverage closer to target. Our ablation study in Figure \ref{fig:merging_set_size} indicates that a modest sample size of $\approx 160$ is sufficient to reduce overcoverage to within $3\%$, while maintaining validity.

\subsection{Evaluation metrics} \label{app:exp_details_evaluation}
\paragraph{Worst-slice (WS) coverage}
We report WS coverage to evaluate the local validity of our prediction sets.

A \textit{slab} of the feature space is a region bounded by hyperplanes:
\begin{equation*}
    S_{v,a,b}=\{x:a \le v^\top x \le b\}.
\end{equation*}
WS coverage is the minimum empirical conditional coverage over all slabs of the feature space that contain at least some fraction $\delta$ of the test points:
\begin{equation*}
    \min_{v,a,b: \mathbb{P}\{X \in S_{v,a,b}\} \ge \delta} \mathbb{P}\{Y \in C(X) | X \in S_{v,a,b}\}. 
\end{equation*}
Intuitively, WS coverage measures performance on the most challenging slice that still has a sufficient number of test samples. In our experiments, we approximate this metric by evaluating coverage over 1000 random projections $v$ and slab boundaries $(a,b)$ and reporting the minimum, following \citet{romano2020classification}.

\subsection{Expert feature assignments for UCI experiments} \label{app:exp_details_uci_features}
For the UCI experiments, the features for each dataset are partitioned into groups of semantically related features, and each expert in the MoE then specializes in a single group of features. We list these groups and their features below, where the feature names are provided with the original dataset \citep{frank2010uci}.

\begin{table}[ht]
\centering
\caption{Airfoil dataset feature groups}
\begin{tabular}{ll}
\toprule
\textbf{Group} & \textbf{Features} \\
\midrule
Aerodynamics   & frequency, free-stream-velocity \\
Geometry       & attack-angle, chord-length, suction-side-displacement-thickness \\
\bottomrule
\end{tabular}
\end{table}

\begin{table}[ht]
\centering
\caption{Wine dataset feature groups}
\begin{tabular}{ll}
\toprule
\textbf{Group} & \textbf{Features} \\
\midrule
Acidity                     & fixed\_acidity, volatile\_acidity, citric\_acid, pH \\
Sugar/alcohol             & residual\_sugar, density, alcohol \\
Sulfur/salinity           & chlorides, free\_sulfur\_dioxide, total\_sulfur\_dioxide, sulphates \\
\bottomrule
\end{tabular}
\end{table}

\begin{table}[ht]
\centering
\caption{Communities dataset feature groups}
\begin{tabularx}{\textwidth}{@{}lX@{}}
\toprule
\textbf{Group} & \textbf{Features} \\
\midrule
Population                  & population, householdsize, numbUrban, pctUrban, LandArea, PopDens, agePct12t21, agePct12t29, agePct16t24, agePct65up \\
Race/ethnicity               & racepctblack, racePctWhite, racePctAsian, racePctHisp, PctForeignBorn, PctSpeakEnglOnly, PctNotSpeakEnglWell, PctBornSameState, PctSameHouse85, PctSameCity85 \\
Income/poverty               & medIncome, medFamInc, perCapInc, whitePerCap, blackPerCap, indianPerCap, AsianPerCap, HispPerCap, NumUnderPov, PctPopUnderPov \\
Employment/industry          & pctWWage, pctWFarmSelf, pctWInvInc, pctWSocSec, pctWPubAsst, pctWRetire, PctUnemployed, PctEmploy, PctEmplManu, PctEmplProfServ \\
Occupation/education         & PctOccupManu, PctOccupMgmtProf, PctWorkMomYoungKids, PctWorkMom, PctUsePubTrans, PctLess9thGrade, PctNotHSGrad, PctBSorMore, MalePctDivorce, MalePctNevMarr \\
Family structure             & FemalePctDiv, TotalPctDiv, PersPerFam, PctFam2Par, PctKids2Par, PctYoungKids2Par, PctTeen2Par, PctLargHouseFam, PctLargHouseOccup, PctSameState85 \\
Housing characteristics      & PersPerOccupHous, PersPerOwnOccHous, PersPerRentOccHous, PctPersOwnOccup, PctHousNoPhone, PctHousLess3BR, MedNumBR, HousVacant, PctHousOccup, PctHousOwnOcc \\
Housing quality         & PctPersDenseHous, PctVacantBoarded, PctVacMore6Mos, MedYrHousBuilt, PctWOFullPlumb, OwnOccLowQuart, OwnOccMedVal, OwnOccHiQuart, RentLowQ, RentMedian \\
Homelessness    & RentHighQ, MedRent, MedRentPctHousInc, MedOwnCostPctInc, MedOwnCostPctIncNoMtg, NumInShelters, NumStreet, NumIlleg, PctIlleg, LemasPctOfficDrugUn \\
Immigration                 & NumImmig, PctImmigRecent, PctImmigRec5, PctImmigRec8, PctImmigRec10, PctRecentImmig, PctRecImmig5, PctRecImmig8, PctRecImmig10 \\
\bottomrule
\end{tabularx}
\end{table}

\begin{table}[ht]
\centering
\caption{Superconductivity dataset feature groups}
\begin{tabularx}{\textwidth}{@{}lX@{}}
\toprule
\textbf{Feature Group} & \textbf{Features} \\
\midrule
Atomic mass            & \seqsplit{mean\_atomic\_mass, ~~wtd\_mean\_atomic\_mass, ~~gmean\_atomic\_mass, ~~wtd\_gmean\_atomic\_mass, ~~entropy\_atomic\_mass, ~~wtd\_entropy\_atomic\_mass, ~~range\_atomic\_mass, ~~wtd\_range\_atomic\_mass, ~~std\_atomic\_mass, ~~wtd\_std\_atomic\_mass, ~~number\_of\_element} \\
Atomic radius          & \seqsplit{mean\_atomic\_radius, ~~wtd\_mean\_atomic\_radius, ~~gmean\_atomic\_radius, ~~wtd\_gmean\_atomic\_radius, ~~entropy\_atomic\_radius, ~~wtd\_entropy\_atomic\_radius, ~~range\_atomic\_radius, ~~wtd\_range\_atomic\_radius, ~~std\_atomic\_radius, ~~wtd\_std\_atomic\_radius} \\
Density                & \seqsplit{mean\_Density, ~~wtd\_mean\_Density, ~~gmean\_Density, ~~wtd\_gmean\_Density, ~~entropy\_Density, ~~wtd\_entropy\_Density, ~~range\_Density, ~~wtd\_range\_Density, ~~std\_Density, ~~wtd\_std\_Density} \\
Electron affinity      & \seqsplit{mean\_ElectronAffinity, ~~wtd\_mean\_ElectronAffinity, ~~gmean\_ElectronAffinity, ~~wtd\_gmean\_ElectronAffinity, ~~entropy\_ElectronAffinity, ~~wtd\_entropy\_ElectronAffinity, ~~range\_ElectronAffinity, ~~wtd\_range\_ElectronAffinity, ~~std\_ElectronAffinity, ~~wtd\_std\_ElectronAffinity} \\
FIE                    & \seqsplit{mean\_fie, ~~wtd\_mean\_fie, ~~gmean\_fie, ~~wtd\_gmean\_fie, ~~entropy\_fie, ~~wtd\_entropy\_fie, ~~range\_fie, ~~wtd\_range\_fie, ~~std\_fie, ~~wtd\_std\_fie} \\
Fusion heat            & \seqsplit{mean\_FusionHeat, ~~wtd\_mean\_FusionHeat, ~~gmean\_FusionHeat, ~~wtd\_gmean\_FusionHeat, ~~entropy\_FusionHeat, ~~wtd\_entropy\_FusionHeat, ~~range\_FusionHeat, ~~wtd\_range\_FusionHeat, ~~std\_FusionHeat, ~~wtd\_std\_FusionHeat} \\
Thermal conductivity   & \seqsplit{mean\_ThermalConductivity, ~~wtd\_mean\_ThermalConductivity, ~~gmean\_ThermalConductivity, ~~wtd\_gmean\_ThermalConductivity, ~~entropy\_ThermalConductivity, ~~wtd\_entropy\_ThermalConductivity, ~~range\_ThermalConductivity, ~~wtd\_range\_ThermalConductivity, ~~std\_ThermalConductivity, ~~wtd\_std\_ThermalConductivity} \\
Valence                & \seqsplit{mean\_Valence, ~~wtd\_mean\_Valence, ~~gmean\_Valence, ~~wtd\_gmean\_Valence, ~~entropy\_Valence, ~~wtd\_entropy\_Valence, ~~range\_Valence, ~~wtd\_range\_Valence, ~~std\_Valence, ~~wtd\_std\_Valence} \\
\bottomrule
\end{tabularx}
\end{table}

\begin{table}[ht]
\centering
\caption{Adult dataset feature groups}
\begin{tabular}{ll}
\toprule
\textbf{Group} & \textbf{Features} \\
\midrule
Demographics         & age, race, sex \\
Education           & education, education-num \\
Economic status     & fnlwgt, capital-gain, capital-loss, hours-per-week \\
Family relationship & marital-status, relationship \\
\bottomrule
\end{tabular}
\end{table}

\begin{table}[ht]
\centering
\caption{Student dataset feature groups}
\begin{tabularx}{\textwidth}{@{}lX@{}}
\toprule
\textbf{Group} & \textbf{Features} \\
\midrule
Personal details     & Marital Status, Gender, Age at enrollment, Nationality, International, Mother's qualification, Father's qualification, Mother's occupation, Father's occupation \\
Academic details     & Application mode, Application order, Previous qualification, Previous qualification (grade), Admission grade, Daytime/evening attendance, Course, Displaced, Educational special needs \\
Performance         & Curricular units 1st sem (credited), Curricular units 1st sem (enrolled), Curricular units 1st sem (evaluations), Curricular units 1st sem (approved), Curricular units 1st sem (grade), Curricular units 1st sem (without evaluations), Curricular units 2nd sem (credited), Curricular units 2nd sem (enrolled), Curricular units 2nd sem (evaluations), Curricular units 2nd sem (approved), Curricular units 2nd sem (grade), Curricular units 2nd sem (without evaluations) \\
Economic and financial & Scholarship holder, Tuition fees up to date, Debtor, Unemployment rate, Inflation rate, GDP \\
\bottomrule
\end{tabularx}
\end{table}

\end{document}